\documentclass{article}

\RequirePackage{silence}
\usepackage[nonatbib, main, final]{neurips_2026}

\usepackage[utf8]{inputenc} 
\usepackage[T1]{fontenc}    
\usepackage{url}         
\usepackage{booktabs}       
\usepackage{amsfonts}      
\usepackage{nicefrac}       
\usepackage{microtype}     
\usepackage[dvipsnames]{xcolor}       

\usepackage{physics}
\usepackage{amsmath}
\usepackage[cjk]{kotex} 
\usepackage{multirow}
\usepackage{booktabs}
\usepackage{tabularx}
\usepackage{colortbl}
\usepackage{bbding}
\usepackage{pifont}
\usepackage{amssymb}
\usepackage{subcaption}
\usepackage{algorithm}
\usepackage{algpseudocodex}
\usepackage{bbm} 
\usepackage[dvipsnames]{xcolor}
\usepackage{wrapfig}
\usepackage[numbers]{natbib}

\definecolor{cvprblue}{rgb}{0.21,0.49,0.74}

\newcommand\han[1]{#1}

\newcommand{\mycc}{\cellcolor{BrickRed!15}}

\newcolumntype{R}{>{\columncolor{gray!20}}c}

\newcommand\clearrow{\global\let\rowmac\relax}
\newcommand{\specialcell}[2][c]{
  \begin{tabular}[#1]{@{}c@{}}#2\end{tabular}}

\newcommand\ours{{\mbox{DBW}}}

\usepackage[pagebackref,breaklinks,colorlinks,hypertexnames=false,citecolor=cvprblue,linkcolor=BrickRed,urlcolor=PineGreen]{hyperref}
\usepackage[capitalize,noabbrev]{cleveref}

\makeatletter
\let\showhyphens\relax
\makeatother

\crefname{section}{Section}{Sections}
\Crefname{section}{Sec.}{Secs.}
\Crefformat{section}{\S#2#1#3}

\newcommand{\tableQuantAcc}{
\begin{table*}[t!]
  \centering
  \footnotesize
  \setlength{\tabcolsep}{2.8pt}
  \renewcommand{\arraystretch}{0.7}
  \begin{tabular}{clccccccccccc}
    \toprule
    {\textbf{LLaMA}}& {\textbf{Method}} & {\textbf{Group}} & {\textbf{Bits}} & {\textbf{PIQA}} & {\textbf{ARC-e}} & {\textbf{HellaS}}& {\textbf{Wing}} & {\textbf{Race}} & {\textbf{ARC-c}} & {\textbf{LAMB-o}} & {\textbf{LAMB-s}} & {\textbf{Avg.}}\\
    \midrule
    \multirow{8}{*}{1-7B} &FP16 & - &16 &78.67 &75.29 &56.99 &70.01 &40.29 &41.81 &73.57 &67.82 & 63.06\\
    \cmidrule{2-13}
    & GPTQ & 128 & 2 & 59.41 & 36.11 & 32.59 & 50.43 & 27.56 & 20.31 & 12.79 & 9.02 & 31.03 \\
    & SliM-LLM\textsuperscript{+} & 128 & 2 & 65.72 & 52.74 & 39.49 & 55.33 & 34.35 & 23.81 & 38.15 & 22.96 & 41.57 \\
     & OmniQuant & 128 & 2 & 67.74 & {59.51} & 41.24 & 56.75 & 33.30 & 28.41 & 40.89 & 25.75 & 44.20 \\
     &PB-LLM & 128 & 1.7(+1) &55.71&29.12&28.31&48.86 &26.41 &19.80 &10.42 &10.09 &28.59 \\
     &BiLLM & 128 & 1(+1.1) &61.10&40.99&31.80 &53.67 &30.14 &20.64 &23.15 &16.48 &36.00\\ 
     &PTQ1.61 & - & 1.61 &63.71&49.62&35.73 &56.75  &32.54  &25.26 &38.93 &26.57 &41.14\\
     & \mycc \ours (Ours) & \mycc 128 & \mycc 1.61 & \mycc {69.53} & \mycc 57.24 & \mycc {41.39} & \mycc {60.38} & \mycc {35.79} & \mycc {28.58} & \mycc {53.74} & \mycc {45.20} & \mycc {48.98} \\
    \midrule
    \multirow{8}{*}{1-13B} &FP16 & - & 16 &79.16 &77.36 &59.92 &72.77 &39.62 & 46.42 &76.15 &71.08 &65.31 \\
    \cmidrule{2-13}
    & GPTQ & 128 & 2 & 65.40 & 44.74 & 38.26 & 56.27 & 33.01 & 24.66 & 26.88 & 15.52 & 38.09 \\
    & SliM-LLM\textsuperscript{+} & 128 & 2 & 71.38 & 60.82 & 44.61 & 62.27 & 36.36 & 30.29 & 50.40 & 33.98 & 48.76 \\
    & OmniQuant & 128 & 2 & 72.20 & {66.58} & {46.86} & {63.69} & {37.42} & {31.83} & 57.21 & 39.39 & 51.90 \\
     &PB-LLM & 128 &1.7(+1) &60.45&37.46&30.79 &51.07 &30.24 &18.69 &27.71 &21.33 &34.72\\
     &BiLLM & 128 &1(+1.1) &67.90 &50.84&39.02 &62.19 & 34.07 &26.11 &49.93 &33.75 &44.98\\
     &PTQ1.61 & - &1.61 &68.17 &58.59 &40.02 &58.33 &34.26 &27.22 &45.95 &35.94 &46.56\\
     & \mycc \ours (Ours) & \mycc 128 & \mycc 1.61 & \mycc {72.25} & \mycc 63.47 & \mycc 46.76 & \mycc 62.67 & \mycc 36.36 & \mycc 31.40 & \mycc {57.60} & \mycc {50.96} & \mycc {52.68} \\
     \midrule
    \multirow{8}{*}{1-30B} &FP16 & - &16 &80.96 &80.39 &63.34 &75.69 &40.57 &52.82 &77.59 &73.34 &68.09\\
    \cmidrule{2-13}
    & GPTQ & 128 & 2 & 68.01 & 55.39 & 43.71 & 60.06 & 35.02 & 27.56 & 45.60 & 27.15 & 45.31 \\
    & SliM-LLM\textsuperscript{+} & 128 & 2 & 72.36 & 65.70 & 49.05 & 64.72 & 37.22 & 33.53 & 60.62 & 44.50 & 53.46 \\
    & OmniQuant & 128 & 2 & 73.12 & 69.40 & 50.71 & 64.25 & 38.66 & 34.64 & 62.95 & 49.89 & 55.45 \\
    &PB-LLM & 128 &1.7(+1) &63.76&40.11&33.32 &61.17 &30.91 &21.33 &43.78 &33.09 &40.93\\
    &BiLLM & 128 & 1(+1.1) & 71.93 & 64.02 & 45.61 & {70.32} & 37.13 & 32.51 & {70.74} & 55.07 & 55.92 \\
    &PTQ1.61 & - & 1.61 &70.24&63.64&46.82 &63.61 &37.13 &32.17 &55.95 &44.61 &51.77\\
    & \mycc \ours (Ours) & \mycc 128 & \mycc 1.61 & \mycc {74.27} & \mycc {70.83} & \mycc {51.81} & \mycc 66.46 & \mycc {40.29} & \mycc {37.71} & \mycc 64.54 & \mycc {56.41} & \mycc {57.79} \\
    \midrule
    \multirow{8}{*}{2-7B}
    & FP16 & - & 16 & 78.07 &76.30 & 57.14 &69.06 &39.52 &43.34 &73.86 &68.23 & 63.19\\
    \cmidrule{2-13}
    & GPTQ & 128 & 2 & 57.29 & 33.42 & 29.42 & 50.67 & 28.71 & 20.65 & 8.05 & 5.82 & 29.25 \\
    & SliM-LLM\textsuperscript{+} & 128 & 2 & 62.68 & 48.44 & 37.19 & 55.09 & 31.77 & 23.38 & 28.97 & 14.30 & 37.73 \\
    & OmniQuant & 128 & 2 & 64.80 & 50.34 & 40.03 & 56.43 & 32.82 & 25.60 & 36.87 & 19.99 & 40.86 \\
     &PB-LLM & 128 &1.7(+1) &54.46&28.20&27.03&49.09&26.70&19.11&7.08&5.71&27.17\\
     &BiLLM & 128 &1(+1.1) &60.39&39.94&30.74&51.93&29.57&21.16&18.44&14.05&33.28\\
     &PTQ1.61 & - & 1.61 &63.22 &47.18 &35.78 &52.25 &29.86 &22.27 &37.38 &25.65 &39.20\\
     & \mycc \ours (Ours) & \mycc 128 & \mycc 1.61 & \mycc {68.61} & \mycc {60.31} & \mycc {41.68} & \mycc {59.27} & \mycc {35.69} & \mycc {27.99} & \mycc {53.72} & \mycc {45.20} & \mycc {49.06} \\
    \midrule
    \multirow{8}{*}{2-13B}
    & FP16 & - &16 &79.11 &79.46 &60.04 &72.14 &40.57 &48.46 &76.77 &70.33 &65.86 \\
    \cmidrule{2-13}
    & GPTQ & 128 & 2 & 61.53 & 40.78 & 34.00 & 53.75 & 29.28 & 19.97 & 20.36 & 13.86 & 34.19 \\
    & SliM-LLM\textsuperscript{+} & 128 & 2 & 65.45 & 55.13 & 39.87 & 55.80 & 34.74 & 25.00 & 33.11 & 20.80 & 41.24 \\
    & OmniQuant & 128 & 2 & 69.31 & 63.51 & 45.11 & 58.01 & 36.84 & 31.57 & 54.05 & 34.37 & 49.10 \\
     &PB-LLM & 128 &1.7(+1) &54.46&27.95&26.74&49.96&26.03&19.54&3.14&2.50&26.29\\
     &BiLLM & 128 &1(+1.1) &63.55 &49.83 &34.36 &58.17 &32.34 &23.81 &40.81 &25.15 &41.00\\
     &PTQ1.61 & - &1.61 &66.54 &56.86 &40.32 &55.88 &33.30 &26.45 &47.23 &31.21 &44.72\\
     & \mycc \ours (Ours) & \mycc 128 & \mycc 1.61 & \mycc {72.31} & \mycc {66.25} & \mycc {47.46} & \mycc {61.80} & \mycc {39.04} & \mycc {33.70} & \mycc {60.64} & \mycc {51.91} & \mycc {54.14} \\
    \midrule
    \multirow{8}{*}{3-8B} 
    &FP16 & - & 16 &79.54 &80.09 &60.14 &73.24 &40.29 &50.17 &75.65 &68.72 &65.98\\
    \cmidrule{2-13}
    & GPTQ & 128 & 2 & 54.95 & 27.36 & 27.48 & 49.72 & 24.21 & 21.50 & 1.22 & 0.52 & 25.87 \\
    & SliM-LLM\textsuperscript{+} & 128 & 2 & 53.26 & 26.35 & 26.43 & 49.49 & 21.15 & 19.37 & 0.00 & 0.00 & 24.51 \\
    & OmniQuant & 128 & 2 & 54.46 & 27.23 & 27.02 & 50.75 & 21.53 & 20.99 & 0.04 & 0.08 & 25.26 \\
     &PB-LLM & 128 &1.7(+1) &56.91&32.37&28.43&49.25&27.66&17.41&16.63&12.44&30.14  \\
     &BiLLM & 128 &1(+1.1) &60.01&38.26&31.48&53.75&30.14&19.45&26.30&15.51&34.36 \\
     &PTQ1.61 & - &1.61 &63.22 &46.17& 34.71 &52.80 &29.09 &23.04 &30.27 &18.26&37.20\\
     & \mycc \ours (Ours) & \mycc 128 & \mycc 1.61 & \mycc {66.38} & \mycc {55.01} & \mycc {37.88} & \mycc {55.96} & \mycc {34.64} & \mycc {25.60} & \mycc {44.19} & \mycc {35.34} & \mycc {44.38} \\
    \bottomrule
  \end{tabular}
    \caption{\textbf{Commonsense reasoning accuracy evaluation.} We report the commonsense reasoning accuracy using the LLaMA model families.
    Following the recent practice in PTQ1.61, we report the bit configuration as weight bits(+mask bits). Specifically, PB-LLM uses 1.7-bit weights with an additional 1-bit mask overhead, while BiLLM employs 1-bit weights with a mask cost of 1.1 bits.}
  \label{tab:zeroshot1}
\end{table*}
}

\newcommand{\tableQuantPPL}{
\begin{table*}[t]
  \centering
  \footnotesize
  \setlength{\tabcolsep}{2.3pt}
  \renewcommand{\arraystretch}{0.9}
  \begin{tabular}{clcccccccccccc}
    \toprule

     {\multirow{2}{*}{{\textbf{Dataset}}}} & 
     \multirow{2}{*}{{{\textbf{Method}}}} & 
     \multirow{2}{*}{{\textbf{Group}}} & 
     \multirow{2}{*}{{{\textbf{Bits}}}} &
     \multicolumn{5}{c}{{\textbf{LLaMA}}} & 
     \multicolumn{4}{c}{{\textbf{LLaMA-2}}} & 
     \multicolumn{1}{c}{{\textbf{LLaMA-3}}} \\
     \cmidrule(lr){5-9}
     \cmidrule(lr){10-13}
     \cmidrule(lr){14-14}
     & & & & \textit{7B}  & \textit{13B} & \textit{30B} & \textit{65B} & \textit{*Avg} & \textit{7B} & \textit{13B} & \textit{70B} & \textit{*Avg} & \textit{8B} \\
    \midrule
    
    \multirow{10}{*}{WikiText2} & FP16 & - & 16 &5.68 &5.09 &4.10 &3.53 & 4.60 &5.47 &4.88 &3.31 & 4.55 &6.14\\
    \cmidrule{2-14}
    & AWQ & 128 & 2 & 2.6e5 & 2.8e5 & 2.4e5 & 7.4e4 & 2.1e5 & 2.2e5 & 1.2e5 & - & 1.7e5 & 1.7e6 \\
    & GPTQ & 128 & 2 & 152.31 & 20.44 & 13.01 & 9.51 & 48.82 & 60.45 & 28.14 & 8.78 & 32.46 & 210.00 \\
    & QuIP & 128 & 2 & 29.74 & 12.48 & 11.57 & 7.83 & 15.41 & 39.73 & 13.48 & 6.64 & 19.95 & 84.97 \\
    & OmniQuant & 128 & 2 & 9.72 & 7.93 & 7.12 & 5.95 & 7.68 & 11.06 & 8.26 & 6.55 & 8.62 & 5.3e3 \\
    & SliM-LLM\textsuperscript{+} & 128 & 2 & 9.68 & {7.17} & 6.41 & 5.74 & 7.25 & 10.87 & 7.59 & 6.44 & 8.30 & 1.3e4 \\
    & PB-LLM & 128 & 1.7(+1) & 102.19 & 48.11 & 26.37 & 12.91 & 47.40 & 66.30 & 462.84 & NaN & 264.57 & 78.67 \\
    & BiLLM & 128 & 1(+1.1) &35.04& 15.14 &10.52 & 8.51 & 17.30 &32.48 &21.77 &12.80 & 22.35 &50.59 \\
    & PTQ1.61 & - & 1.61 & 12.50 & 9.67 & 7.95 & 7.02 & 9.29 & 12.70 &9.74 &6.94 & 9.79 &22.90 \\
    & \mycc \ours (Ours) & \mycc 128 & \mycc 1.61 & \mycc {9.13} & \mycc 7.65 & \mycc {6.40} & \mycc {5.57} & \mycc {7.19} & \mycc {8.78} & \mycc {7.37} & \mycc {6.10} & \mycc {7.42} & \mycc {15.96} \\
    \midrule
    
    \multirow{10}{*}{C4}&FP16 & - & 16 &7.08 &6.61 &5.98 &5.62 & 6.32 &6.97 &6.46 &5.52 & 6.32 &8.88\\
    \cmidrule{2-14}
    & AWQ & 128 & 2 & 1.9e5 & 2.3e5 & 2.4e5 & 7.5e4 & 1.8e5 & 1.7e5 & 9.4e4 & - & 1.3e5 & 2.1e6 \\
    & GPTQ & 128 & 2 & 34.63 & 15.29 & 11.93 & 11.99 & 18.46 & 33.70 & 20.97 & NaN & 27.34 & 4.1e4 \\
    & QuIP & 128 & 2 & 33.74 & 21.94 & 10.95 & 13.99 & 20.16 & 31.94 & 16.16 & 8.17 & 18.76 & 1.3e2 \\
    & OmniQuant & 128 & 2 & 12.97 & 10.36 & 9.36 & 8.00 & 10.17 & 15.02 & 11.05 & 8.52 & 11.53 & 6.1e3 \\
    & SliM-LLM\textsuperscript{+} & 128 & 2 & 14.99 & 10.22 & 9.33 & {7.52} & 10.52 & 18.18 & {10.24} & 8.40 & 12.27 & 1.7e4 \\
    & PB-LLM & 128 & 1.7(+1)  &67.92& 34.20 &22.45 & 13.70 & 34.57 & 66.23 &333.54 & NaN & 199.89 & 78.98 \\
    & BiLLM & 128 & 1(+1.1) &33.64& 14.75 & 10.97 & 9.92 & 17.32 &33.72 &23.14 &13.84 & 23.57 & 48.26 \\
    & PTQ1.61 & - &1.61 &17.13 & 13.51 & 10.98 & 9.86 & 12.87 & 17.73 &13.64 &12.63 & 14.66 & 33.82 \\
    & \mycc \ours (Ours) & \mycc 128 & \mycc 1.61 & \mycc {12.16} & \mycc {10.17} & \mycc {8.93} & \mycc 7.92 & \mycc {9.80} & \mycc {12.30} & \mycc 10.42 & \mycc {8.15} & \mycc {10.29} & \mycc {22.79} \\
    \bottomrule
    
  \end{tabular}
  \caption{\textbf{Perplexity evaluation.} We report perplexity on WikiText2 and C4 across LLaMA models. 
  \textit{*Avg} is the arithmetic mean over available model sizes, excluding missing (-) or NaN entries.}
  \label{tab:ppl}
\end{table*}
}

\newcommand{\tableBoth}{
\begin{figure*}[t]
\centering
\begin{minipage}[t]{0.43\linewidth}
\vspace{0pt}
\centering
\footnotesize
\setlength{\tabcolsep}{4.2pt}
\renewcommand{\arraystretch}{0.96}

\begin{tabular}{clccc}
\toprule
\textbf{Bit} & \textbf{Method} & \textbf{Group} & \textbf{PPL\textcolor{cvprblue}{↓}} & \textbf{Acc.\textcolor{cvprblue}{↑}} \\
\midrule

16 & FP16 & - & 6.22 & 63.19 \\
\midrule

\multirow{3}{*}{2} & 
OmniQuant & 128 & 13.04 & 40.86 \\
& SliM-LLM\textsuperscript{+} & 128 & 14.53 & 37.73 \\
& \mycc \ours (Ours) & \mycc 128 & \mycc 9.56 & \mycc 51.90 \\
\midrule

1(+1.1) &  
BiLLM & 128 & 33.10 & 33.28 \\
1.61 & PTQ1.61 & - & 15.22 & 39.20 \\
1.61 & \mycc \ours (Ours) & \mycc 128 & \mycc 10.54 & \mycc 49.06 \\
\midrule
1 & \mycc \ours (Ours) & \mycc 128 & \mycc 14.97 & \mycc 39.66 \\
0.5 & \mycc \ours (Ours) & \mycc 128 & \mycc 27.68 & \mycc 30.93 \\
\bottomrule
\end{tabular}

\captionof{table}{\textbf{Bit-width evaluation.} 
We evaluate PTQ baselines on various bit-widths using the LLaMA-2-7B model.}
\label{tab:bitwidth}

\end{minipage}
\hfill
\begin{minipage}[t]{0.55\linewidth}
\vspace{0pt}
\centering

\footnotesize
\setlength{\tabcolsep}{3.9pt}
\renewcommand{\arraystretch}{0.85}

\begin{tabular}{clcccc}
\toprule
\textbf{Model} & \textbf{Method} & \textbf{Bit} & \textbf{Group} & \textbf{PPL\textcolor{cvprblue}{↓}} & \textbf{Acc.\textcolor{cvprblue}{↑}} \\
\midrule

\multirow{3}{*}{Gemma-7B} & FP16 & 16 & - & 8.82 & 66.37 \\ 
\cmidrule{2-6}
~ & BiLLM & 1(+1.1) & 128 & 5.9e13 & 24.77 \\ 
~ & \mycc{\ours (Ours)} & \mycc{1.61} & \mycc{128} & \mycc{74.49} & \mycc{30.27} \\ 
\midrule
\multirow{3}{*}{Qwen-2.5-7B} & FP16 & 16 & - & 6.82 & 66.65 \\ 
\cmidrule{2-6}
~ & BiLLM & 1(+1.1) & 128 & 33.86 & 43.88 \\ 
~ & \mycc{\ours (Ours)} & \mycc{1.61} & \mycc{128} & \mycc{20.98} & \mycc{44.61} \\
\midrule
\multirow{3}{*}{Mistral-7B} & FP16 & 16 & - & 9.37 & 64.80 \\ 
\cmidrule{2-6}
~ & BiLLM & 1(+1.1) & 128 & 70.55 & 28.49 \\ 
~ & \mycc{\ours (Ours)} & \mycc{1.61} & \mycc{128} & \mycc{14.30} & \mycc{45.64} \\ 
\bottomrule
\end{tabular}

\captionof{table}{\textbf{Non-LLaMA model evaluation.} We compare ours against the BiLLM, an advanced sub-2-bit PTQ method generalizing well on different models.}
\label{tab:model}

\end{minipage}

\end{figure*}
}

\newcommand{\tableSVD}{
\begin{table*}[t]
    \centering
    \footnotesize
    \setlength{\tabcolsep}{2pt}
    \renewcommand{\arraystretch}{0.9}
    \begin{tabular}{clcccccccccccc}
    \toprule
    \multirow{2}{*}{{\textbf{Type}}} &
    \multirow{2}{*}{{\textbf{Method (Bit)}}} & 
    \multicolumn{4}{c}{{\textbf{LLaMA-7B}}} &
    \multicolumn{4}{c}{{\textbf{LLaMA-2-7B}}} &
    \multicolumn{4}{c}{{\textbf{LLaMA-3-8B}}} \\
    \cmidrule(lr){3-6}
    \cmidrule(lr){7-10}
    \cmidrule(lr){11-14}
    
    & &
    \textit{Ratio} &  
    \textit{Mem} & 
    \textit{PPL\textcolor{cvprblue}{↓}} & \textit{Acc.\textcolor{cvprblue}{↑}} &
    
    \textit{Ratio} & 
    \textit{Mem} & 
    \textit{PPL\textcolor{cvprblue}{↓}} & \textit{Acc.\textcolor{cvprblue}{↑}} &
    
    \textit{Ratio} & 
    \textit{Mem} & 
    \textit{PPL\textcolor{cvprblue}{↓}} & \textit{Acc.\textcolor{cvprblue}{↑}} \\
    \midrule
    
    - & Baseline (FP16) & 
    0.00 & 
    12.31 & 6.38 & 63.06 &
    0.00 & 
    12.31 & 6.22 & 63.19 &
    0.00 &
    13.98 & 7.51 & 65.98 \\
    \midrule

    \multirow{6}{*}{\rotatebox[origin=c]{90}{\specialcell{sequential\\optimization}}} 
    & ASVD (FP16) & 
    46.33 &
    6.72 & 3.0e4 & 25.02 &
    46.33 &
    6.72 & 4.5e4 & 23.94 &
    46.33 & 
    7.96 & 2.6e4 & 24.43 \\
    & ASVD+GPTQ (1.61) & 
    46.33 &
    1.56 & 3.0e4 & 25.00 &
    46.33 & 
    1.56 & 4.5e4 & 24.63 &
    46.33 & 
    2.40 & 2.6e4 & 24.26 \\
    & ASVD+OmniQ (1.61) & 
    46.33 & 
    1.56 & 3.1e4 & 25.02 &
    46.33 & 
    1.56 & 8.9e4 & 24.89 &
    46.33 & 
    2.40 & 5.3e4 & 24.97 \\
    \cmidrule(lr){2-14}
    
    & SVD-LLM (FP16) & 
    46.33 & 
    6.72 & 14.08 & 47.21 &
    46.33 & 
    6.72 & 59.28 & 31.59 &
    46.33 & 
    7.96 & 53.76 & 36.55 \\
    & SVD-LLM+GPTQ (1.61) & 
    46.33 & 
    1.56 & 22.45 & 37.17 &
    46.33 & 
    1.56 & 116.38 & 27.80 &
    46.33 & 
    2.40 & 102.31 & 30.26 \\
    & SVD-LLM+OmniQ (1.61) & 
    46.33 & 
    1.56 & 17.22 & 42.84 &
    46.33 &
    1.56 & 77.10 & 29.35 &
    46.33 & 
    2.40 & 89.41 & 30.52 \\
    \midrule

    \multicolumn{2}{l}{\mycc co-optimization (our \ours) (1.61)} & 
    \mycc 47.21 & 
    \mycc 1.56 & \mycc 10.65 & \mycc 48.98 &
    \mycc 46.95 & 
    \mycc 1.56 & \mycc 10.54 & \mycc 49.06 &
    \mycc 44.71 & 
    \mycc 2.40 & \mycc 19.38 & \mycc 44.38 \\
    \bottomrule
    \end{tabular}
    \caption{\textbf{Sequential vs. co-optimization of pruning and quantization.} 
    We evaluate compression methods using SVD with PTQ approaches.
    ASVD and SVD-LLM apply the sequential optimization of pruning and quantization, while ours co-optimizes them concurrently. Note that methods in parentheses with FP16 denote the pruning-only results.
    We report the average perplexity (\textit{PPL\textcolor{cvprblue}{↓}}) of the WikiText-2 and C4 datasets and the mean accuracy (\textit{Acc.\textcolor{cvprblue}{↑}}) on the eight commonsense benchmarks. 
    \textit{Mem} is the memory footprint (GB) of the compressed model. 
    For a controlled comparison, we match the pruning \textit{Ratio} and bit-width of the baselines to those of \ours, providing more results in~\cref{sec:appendix_sequential_hyperparameter_search}.
    }
    \label{tab:svd_quant}
\end{table*}
}

\newcommand{\tablepruning}{
\begin{table*}[t]
    \centering
    \footnotesize
    \setlength{\tabcolsep}{4pt}
    \renewcommand{\arraystretch}{0.9}
    \begin{tabular}{lcccccccccccc}
    \toprule
    \multirow{2}{*}{{\textbf{Method (Bit)}}} & 
    \multicolumn{4}{c}{{\textbf{LLaMA-7B}}} &
    \multicolumn{4}{c}{{\textbf{LLaMA-2-7B}}} &
    \multicolumn{4}{c}{{\textbf{LLaMA-3-8B}}} \\
    \cmidrule(lr){2-5}
    \cmidrule(lr){6-9}
    \cmidrule(lr){10-13}
    
    &
    \textit{Ratio} & \textit{Mem} & 
    \textit{PPL\textcolor{cvprblue}{↓}} & \textit{Acc.\textcolor{cvprblue}{↑}} &

    \textit{Ratio} & \textit{Mem} & 
    \textit{PPL\textcolor{cvprblue}{↓}} & \textit{Acc.\textcolor{cvprblue}{↑}} &
 
    \textit{Ratio} & \textit{Mem} & 
    \textit{PPL\textcolor{cvprblue}{↓}} & \textit{Acc.\textcolor{cvprblue}{↑}} \\
    \midrule
    Baseline (FP16) & 
    
    0.00 & 12.31 & 6.38 & 63.06 &
    0.00 & 12.31 & 6.22 & 63.19 &
    0.00 & 13.98 & 7.51 & 65.98 \\
    \midrule
    
    LLM-Pruner (FP16) & 
    44.40 & 6.84 & 96.46 & 39.70 &
    44.40 & 6.84 & 108.33 & 38.43 &
    43.50 & 7.90 & 93.37 & 41.45 \\
    
    StreamLine (FP16) & 
    45.10 & 6.76 & 2.0e3 & 29.98 &
    45.10 & 6.76 & 126.99 & 35.85 &
    43.50 & 7.90 & 63.66 & 36.85 \\
    \midrule
    
    \mycc Our \ours~(1.61) & 
    \mycc 47.21 & \mycc 1.56 & \mycc 10.65 & \mycc 48.98 &
    \mycc 46.95 & \mycc 1.56 & \mycc 10.54 & \mycc 49.06 &
    \mycc 44.71 & \mycc 2.40 & \mycc 19.38 & \mycc 44.38 \\
    \bottomrule
    \end{tabular}
    \caption{
    \textbf{Structured pruning method evaluation under matched pruning ratio.}
    We compare DBW with structured pruning baselines by matching their pruning \textit{Ratios}. The structured pruning baselines keep the remaining weights in FP16, whereas DBW combines SVD-rank pruning with low-bit quantization; therefore, \textit{Mem} is reported to make the storage difference explicit.
    }
    \label{tab:pruning}
\end{table*}
}

\newcommand{\tableFactorAnalysis}{
\begin{table*}[t]
\centering
\footnotesize
\setlength{\tabcolsep}{4pt}
\renewcommand{\arraystretch}{0.9}

\begin{subtable}{0.5\textwidth}
    \centering
    \begin{tabular}{clcc}
    \toprule
    \textbf{LLaMA} & \textbf{Method} & \textbf{PPL\textcolor{cvprblue}{↓}} & \textbf{Acc.\textcolor{cvprblue}{↑}} \\
    \midrule
    
    \multirow{3}{*}{1-7B} 
    & Uniform 
    & 690.8\;\;\;\;\;\;\;\;\;\;\;\;\;\; & 25.06\;\;\;\;\;\;\;\;\;\;\;\;\;\; \\ 
    
    & Heuristic 
    & 11.71\;(\textcolor{cvprblue}{-679.09}) & 45.54\;(\textcolor{cvprblue}{+20.48}) \\ 
    
    & \mycc{Learnable} 
    & \mycc{10.65\;(\textcolor{cvprblue}{-680.15})} 
    & \mycc{48.98\;(\textcolor{cvprblue}{+23.92})} \\ 
    
    \midrule
    
    \multirow{3}{*}{2-7B} 
    & Uniform 
    & 650.81\;\;\;\;\;\;\;\;\;\;\;\;\;\;\;\; & 25.05\;\;\;\;\;\;\;\;\;\;\;\;\;\; \\ 
    
    & Heuristic 
    & 11.69\;(\textcolor{cvprblue}{-639.12}) 
    & 45.10\;(\textcolor{cvprblue}{+20.05}) \\ 
    
    & \mycc{Learnable} 
    & \mycc{10.54\;(\textcolor{cvprblue}{-640.27})} 
    & \mycc{49.06\;(\textcolor{cvprblue}{+24.01})} \\ 
    
    \bottomrule
    \end{tabular}
    \caption{Bit-allocation method}
    \label{tab:factor1}
\end{subtable}
\begin{subtable}{0.45\textwidth}
    \centering
    \begin{tabular}{clcc}
    \toprule
    \textbf{LLaMA} & \textbf{Target} & \textbf{PPL\textcolor{cvprblue}{↓}} & \textbf{Acc.\textcolor{cvprblue}{↑}} \\
    \midrule
    
    \multirow{3}{*}{1-7B}
    & $\beta$ 
    & 10.90\;\;\;\;\;\;\;\;\;\;\; & 47.25\;\;\;\;\;\;\;\;\;\;\;\; \\ 
    & $\beta,\alpha$ 
    & 10.54\;(\textcolor{cvprblue}{-0.36}) 
    & 47.55\;(\textcolor{cvprblue}{+0.30}) \\ 
    & \mycc{$\beta,\alpha,\sigma$} 
    & \mycc{10.65\;(\textcolor{cvprblue}{-0.25})} 
    & \mycc{48.98\;(\textcolor{cvprblue}{+1.73})} \\ 
    
    \midrule
    
    \multirow{3}{*}{2-7B}
    & $\beta$ 
    & 11.01\;\;\;\;\;\;\;\;\;\;\; & 48.42\;\;\;\;\;\;\;\;\;\;\;\; \\ 
    & $\beta,\alpha$ 
    & 10.74\;(\textcolor{cvprblue}{-0.27}) 
    & 48.80\;(\textcolor{cvprblue}{+0.38}) \\ 
    & \mycc{$\beta,\alpha,\sigma$} 
    & \mycc{10.54\;(\textcolor{cvprblue}{-0.47})} 
    & \mycc{49.06\;(\textcolor{cvprblue}{+0.64})} \\ 
    
    \bottomrule
    \end{tabular}
    \caption{Learnable target}
    \label{tab:factor2}
\end{subtable}

\caption{\textbf{Controlled factor analysis of SVD-space bit allocation.} 
We disentangle two factors of DBW: (a) the bit-allocation method and (b) the set of learnable targets. 
All variants are evaluated under the same training protocol; 
only the factor specified in each sub-table is changed. In (a), \emph{Uniform} denotes a non-adaptive allocation that assigns a fixed precision pattern to SVD components, while \emph{Heuristic} denotes a non-differentiable SVD-aware mixed-bit allocation based on singular-value importance under the same budget. \emph{Learnable} replaces this static/discrete allocation with our differentiable bit-width learning. In (b), we progressively enable learning of component-level bit-widths $\beta$, matrix-level bit budgets $\alpha$, and singular values $\sigma$. PPL is the average perplexity on WikiText2 and C4, and Acc. is the average accuracy on eight commonsense reasoning benchmarks.
}
\label{tab:factor}
\end{table*}
}

\newcommand{\tableHyperparameter}{
\begin{table*}[h]
\centering
\footnotesize
\setlength{\tabcolsep}{6pt}
\renewcommand{\arraystretch}{0.92}
\begin{tabular}{lc}
\toprule
\textbf{Type} & \textbf{Value} \\
\midrule
dataset & RedPajama-1T \\
\#samples & 4096 \\
sequence length & 2048 \\
epoch & 2 \\
optimizer & adam \\
lr scheduler & cosine \\
lr (weight, comp. bit) & 2e-5 \\
lr (matrix bit) & 1e-4 \\
\bottomrule
\end{tabular}
\caption{\textbf{Hyperparameter specification.} 
We provide an in-depth hyperparameter specification.}
\label{tab:hyperparameter}
\end{table*}
}

\newcommand{\tablecalibration}{
\begin{table*}[h]
\centering
\footnotesize
\setlength{\tabcolsep}{6pt}
\renewcommand{\arraystretch}{0.92}
\begin{tabular}{lccc}
\toprule
\textbf{Method} & \textbf{Dataset} & \textbf{\#SeenSamples} \\
\midrule
SVD-LLM & Alpaca & 50,000 \\
OmniQuant & WikiText-2 & 5,120 \\ 
SliM-LLM\textsuperscript{+} & WikiText-2 & 6,400 \\ 
PTQ1.61 & RedPajama-1T & 20,000 \\
\ours (Ours) & RedPajama-1T & 8,192\\
\bottomrule

\end{tabular}
\caption{\textbf{Calibration setting comparison.} 
We compare our calibration setting against the existing SVD pruning and post-training quantization method. For the methods using multiple datasets, we only report the main dataset that requires the most time and GPU memory. The \#SeenSamples denotes the total number of samples seen during training, which is computed as the number of epochs multiplied by the number of samples.}
\label{tab:calibration}
\end{table*}
}

\newcommand{\tablestat}{
\begin{table*}[h]
\centering
\footnotesize
\setlength{\tabcolsep}{3pt}
\renewcommand{\arraystretch}{0.9}
    \begin{tabular}{lccc}
        \toprule
        \textbf{Metric} &
        \textbf{Mean} &
        \textbf{Std.} &
        \textbf{95\% CI} \\
        \midrule
        PPL $\downarrow$  & 10.58 & $\pm$0.04 & $\pm$0.09 \\
        Acc. $\uparrow$   & 49.32 & $\pm$0.23 & $\pm$0.57 \\
        \bottomrule
    \end{tabular}
    \caption{\textbf{Statistical variation analysis.} We report statistical values of LLaMA-2-7B models over three calibration seeds such as 44, 45, and 46.}
    \label{tab:statistical_variation}
\end{table*}
}

\newcommand{\tableHamilton}{
\begin{table*}[h]
\centering
\footnotesize
\setlength{\tabcolsep}{6pt}
\renewcommand{\arraystretch}{0.92}
\begin{tabular}{lccc}
\toprule
\textbf{Method} & \textbf{Leftover bit allocation} & \textbf{PPL\textcolor{cvprblue}{↓}} & \textbf{Acc.\textcolor{cvprblue}{↑}} \\
\midrule
Floor only & $h_i=0$ & 11.83 & 46.13 \\
Random & $ h_i =
\begin{cases}
1, & i \in \gamma_{\mathrm{rand}}, \\
0, & i \notin \gamma_{\mathrm{rand}},
\end{cases}$ & 11.86 & 46.18 \\
Hamilton & 
$h_i = \begin{cases} 1, & i \in \gamma_{\mathrm{ham}}, \\ 0, & i \notin \gamma_{\mathrm{Ham}}. \end{cases}$ & 
10.54 & 49.06 \\
\bottomrule
\end{tabular}
\caption{\textbf{Hamilton correction analysis.} 
We compare different leftover bit allocation methods for assigning the leftover bit indicator $h_i$ after flooring the continuous bit allocation. $\gamma_{\mathrm{rand}}$ is a uniformly sampled subset of feasible components with $|\gamma_{\mathrm{rand}}|=R$. $\gamma_{\mathrm{ham}}$ denotes the top-$R$ components with the largest remainders. Floor only ignores the leftover bits, Random assigns them to random feasible components, and Hamilton uses the largest-remainder rule. Hamilton gives a deterministic budget-preserving projection and yields stable final performance. This analysis uses the LLaMA-2-7B model.}
\label{tab:hamilton}
\end{table*}
}

\newcommand{\tablebitpack}{
\begin{table*}[t]
\centering
\footnotesize
\setlength{\tabcolsep}{4.pt}
\renewcommand{\arraystretch}{0.92}
\begin{tabular}{lccccc}
\toprule
\textbf{Bit} & \textbf{Memory} & \textbf{Compression} & \textbf{Runtime} & \textbf{Speed up} \\ 
\midrule
FP16 & 1,544MB & - & 3.08ms & - \\
\midrule
INT5 & 483MB & $3.2\times$ & 2.87ms & $1.1\times$ \\
INT4 & 386MB & $4.0\times$ & 2.24ms & $1.4\times$ \\
INT3 & 290MB & $5.3\times$ & 2.42ms & $1.3\times$ \\
INT2 & 193MB & $8.0\times$ & 1.83ms & $1.7\times$ \\
INT1 & 97MB & $16.0\times$ & 1.72ms & $1.8\times$ \\
\bottomrule
\end{tabular}
\caption{\textbf{Our bit-wise GEMM analysis.} 
We report the total weight memory footprint (MB) of one LLaMA-65B block, which consists of seven matrices (\textit{e.g.}, $8192\times8192$ or $8192\times22016$) and measure the runtime of GEMM operation (ms) during generation tasks, across sequence lengths ranging from 1 to 128. For stable results, the runtime is the average values of 30 iterations with a 10-warm-up. 
We demonstrate that the bit-width non-evenly packable to 32-bit word (\textit{e.g.}, 5, 3-bit) can achieve the theoretical compression ratio as well as achieve fast GEMM latency.}
\label{tab:tablebitpack}
\end{table*}
}

\newcommand{\tablegroupsize}{
\begin{table*}[h]
\centering
\footnotesize
\setlength{\tabcolsep}{6pt}
\renewcommand{\arraystretch}{0.9}
    \begin{tabular}{ccccc}
        \toprule
        \textbf{Group} & 
        \textbf{Method} & 
        \textbf{Bit} & 
        \textbf{PPL\textcolor{cvprblue}{↓}} & 
        \textbf{Acc.\textcolor{cvprblue}{↑}} \\
        \midrule
        --  & FP16     & 16        & 6.22  & 63.19 \\
        \midrule
        128 & BiLLM    & 1(+1.1)   & 33.10 & 33.28 \\
        128 & DBW(Ours) & 1.61 & 10.54 & 49.06 \\
        \midrule
        256 & BiLLM    & 1(+1.1)   & 39.55 & 29.65 \\
        256 & DBW(Ours) & 1.61 & 11.02 & 47.72 \\
        \bottomrule
    \end{tabular}
    \caption{
    \textbf{Group size analysis.}
    We evaluate the DBW and BiLLM with different group sizes in terms of perplexity and accuracy. We apply the quantization on the LLaMA-2-7B model.
    }
    \label{tab:group_size}
\end{table*}
}

\newcommand{\tableinstruct}{
\begin{table*}[t]
\centering
\footnotesize
\setlength{\tabcolsep}{3pt}
\renewcommand{\arraystretch}{0.9}
    \begin{tabular}{llcccc}
        \toprule
        \textbf{Model} &
        \textbf{Method} &
        \textbf{Bit} &
        \textbf{Group} &
        \textbf{PPL\textcolor{cvprblue}{↓}} & 
        \textbf{Acc.\textcolor{cvprblue}{↑}} \\
        \midrule

        \multirow{5}{*}{LLaMA-3.1-8B-Instruct}
        & FP16      & 16      & --  & 9.30    & 66.37 \\
        & OmniQuant & 2       & 128 & 112.38  & 27.19 \\
        & SliM-LLM\textsuperscript{+} & 2       & 128 & 301.27  & 26.08 \\
        & BiLLM     & 1(+1.1) & 128 & 51.55   & 34.97 \\
        & DBW(Ours)   & 1.61    & 128 & 18.01   & 45.95 \\
        \midrule

        \multirow{5}{*}{DeepSeek-R1-Distill-LLaMA-8B}
        & FP16      & 16      & --  & 17.21           & 58.09 \\
        & OmniQuant & 2       & 128 & 2.7e3 & 25.32 \\
        & SliM-LLM\textsuperscript{+} & 2       & 128 & 1.1e4 & 24.92 \\
        & BiLLM     & 1(+1.1) & 128 & 1.7e2 & 29.36 \\
        & DBW(Ours)   & 1.61    & 128 & 21.10           & 45.02 \\
        \midrule

        \multirow{5}{*}{LLaMA-2-7B-chat}
        & FP16      & 16      & --  & 8.23   & 62.46 \\
        & OmniQuant & 2       & 128 & 21.23  & 36.25 \\
        & SliM-LLM\textsuperscript{+} & 2       & 128 & 29.41  & 32.20 \\
        & BiLLM     & 1(+1.1) & 128 & 34.46  & 35.88 \\
        & DBW(Ours)   & 1.61    & 128 & 13.27  & 43.42 \\
        \bottomrule
    \end{tabular}
    \caption{
    \textbf{Instruction-tuned model evaluation.}
    This analysis assesses the perplexity and accuracy of the sub-2-bit quantization methods on the instruction-tuned LLMs. Our DBW works well on the post-tuned LLMs.
    }
    \label{tab:instruct_models}
\end{table*}
}

\newcommand{\tableQuantAccTwo}{
\begin{table*}[t]
    \centering
    \footnotesize
    \setlength{\tabcolsep}{6pt}
    \renewcommand{\arraystretch}{0.92}
    \begin{tabular}{lcccccc}
    \toprule
    \multirow{2}{*}{\textbf{Method}} & 
    \multirow{2}{*}{\textbf{Group}} &
    \multirow{2}{*}{\textbf{Bits}} &
    \multicolumn{2}{c}{\textbf{LLaMA-7B}} &
    \multicolumn{2}{c}{\textbf{LLaMA-2-7B}}\\
    \cmidrule(lr){4-5}
    \cmidrule(lr){6-7}
    
    & & &
    \textit{MMLU} & \textit{GSM8K} &
    \textit{MMLU} & \textit{GSM8K}
    \\
    \midrule
    PB-LLM & 128 & 1.7(+1) & 
    23.00 & 0.23 &
    23.00 & 0.00 \\
    BiLLM & 128 & 1(+1.1) & 
    22.90 & 0.00 &
    22.90 & 0.00 \\
    PTQ1.61 & - & 1.61 & 
    23.00 & 0.15 &
    22.90 & 0.61 \\
    \ours (Ours) & 128 & 1.61 & 
    24.47 & 1.82 & 
    23.89 & 2.12 
    \\
    
    \bottomrule
    \end{tabular}
    \caption{\textbf{PTQ knowledge-mathematical reasoning accuracy evaluation.} We evaluate zero-shot reasoning accuracy for the ultra-efficient method on the MMLU and GSM8K datasets. The numbers of PB-LLM and BiLLM come from PTQ1.61.}
    \label{tab:zeroshot2}
\end{table*}
}

\newcommand{\tableSequentialSearch}{
\begin{table*}[t]
\centering
\footnotesize
\setlength{\tabcolsep}{6pt}
\renewcommand{\arraystretch}{0.92}
\begin{tabular}{clcccccc}
\toprule
\multirow{2}{*}{\textbf{Type}} & \multirow{2}{*}{\textbf{Method}} & 
\multicolumn{3}{c}{\textbf{Trade-off}} & 
\multicolumn{3}{c}{\textbf{Perplexity}} \\
\cmidrule(lr){3-5}
\cmidrule(lr){6-8}
& & \textit{Ratio} & \textit{R-bit} & \textit{Bit} & 
\textit{WikiText2} & \textit{C4} & \textit{Avg.} \\ 
\midrule
\multirow{4}{*}{\rotatebox[origin=c]{90}{\specialcell{sequential\\optim.}}} & 
SVD-LLM+AWQ & 20.00 & 4.00 & 3.20 & 7.97 & 11.71 & 9.84 \\ 
& SVD-LLM+GPTQ & 20.00 & 4.00 & 3.20 & 8.14 & 11.45 & 9.79 \\ 
& SVD-LLM+AWQ & 20.00 & 3.00 & 2.40 & 12.47 & 18.57 & 15.52 \\ 
& SVD-LLM+GPTQ & 20.00 & 3.00 & 2.40 & 14.47 & 15.42 & 14.94 \\ 
\midrule
\rowcolor{BrickRed!20}
\multicolumn{2}{l}{{co-optimization (our \ours)}} & 35.54 & 3.10 & 2.00 & 8.27 & 10.85 & \textbf{9.56} \\ 
\midrule

\multirow{4}{*}{\rotatebox[origin=c]{90}{\specialcell{sequential\\optim.}}} &
SVD-LLM+AWQ & 40.00 & 4.00 & 1.92 & 10.24 & 16.58 & 13.41 \\ 
& SVD-LLM+GPTQ & 40.00 & 4.00 & 1.92 & 10.35 & 16.08 & 13.22 \\ 
& SVD-LLM+AWQ & 20.00 & 2.00 & 1.60 & 2.7e5 & 2.5e5 & 2.6e5 \\ 
& SVD-LLM+GPTQ & 20.00 & 2.00 & 1.60 & 6.3e4 & 2.3e4 & 4.3e4 \\ 
\midrule
\rowcolor{BrickRed!20}
\multicolumn{2}{l}{{co-optimization (our \ours)}} & 47.21 & 3.05 & 1.61 & 8.78 & 12.30 & \textbf{10.54} \\ 
\midrule

\multirow{4}{*}{\rotatebox[origin=c]{90}{\specialcell{sequential\\optim.}}} &
SVD-LLM+AWQ & 40.00 & 3.00 & 1.44 & 16.12 & 26.70 & 21.41 \\ 
& SVD-LLM+GPTQ & 40.00 & 3.00 & 1.44 & 16.60 & 21.02 & 18.81 \\ 
& SVD-LLM+AWQ & 40.00 & 2.00 & 0.96 & 2.8e5 & 2.6e5 & 2.7e5 \\ 
& SVD-LLM+GPTQ & 40.00 & 2.00 & 0.96 & 6.8e4 & 2.2e4 & 4.5e4 \\ 
\midrule
\rowcolor{BrickRed!20}
\multicolumn{2}{l}{{co-optimization (our \ours)}} & 65.08 & 2.86 & 1.00 & 12.62 & 17.32 & \textbf{14.97} \\ 
\midrule

\multirow{4}{*}{\rotatebox[origin=c]{90}{\specialcell{sequential\\optim.}}} &
SVD-LLM+AWQ & 60.00 & 4.00 & 0.77 & 16.63 & 28.65 & 22.64 \\ 
& SVD-LLM+GPTQ & 60.00 & 4.00 & 0.77 & 16.26 & 26.97 & \textbf{21.62} \\ 
& SVD-LLM+AWQ & 60.00 & 3.00 & 0.58 & 39.41 & 56.84 & 48.13 \\ 
& SVD-LLM+GPTQ & 60.00 & 3.00 & 0.58 & 25.27 & 35.91 & 30.59 \\ 
\midrule
\rowcolor{BrickRed!20}
\multicolumn{2}{l}{{co-optimization (our \ours)}} & 82.25 & 2.82 & 0.50 & 22.89 & 32.47 & {27.68} \\ 
\bottomrule
\end{tabular}
\caption{\textbf{Extended sequential vs. co-optimization evaluation.} 
We compare sequential vs. co-optimization on different bit-widths. We use the LLaMA-2-7B model. The sequential optimization method employs SVD-LLM for pruning and AWQ and GPTQ for quantizing the remaining components. For a comprehensive evaluation, we attempt to vary pruning \textit{ratio}, bit-width of remaining components (\textit{R-bit}), and quantization method of the sequential optimization method. The proposed co-optimization method works well against the sequential optimization baseline on different bit-widths.}
\label{tab:tableSequentialSearch}
\end{table*}
}

\newcommand{\tabletradeoff}{
\begin{table*}[t]
\centering
\footnotesize
\setlength{\tabcolsep}{6pt}
\renewcommand{\arraystretch}{0.92}
\begin{tabular}{lcccccc}
    \toprule
    \multirow{2}{*}{\textbf{LLaMA}} & \multicolumn{3}{c}{\textbf{Trade-off}} & \multicolumn{3}{c}{\textbf{Weight memory (GB)}} \\
    \cmidrule(lr){2-4}
    \cmidrule(lr){5-7}
    & \textit{Ratio} & \textit{R-bit} & \textit{Bit} & \textit{FP16} & \textit{Ours} & \textit{Size ↓} \\ 
    \midrule
    1-7B & 47.21 & 3.05 & 1.61 & 12.31 & 1.56 & $7.89\times$ \\ 
    1-13B & 47.99 & 3.10 & 1.61 & 23.94 & 2.88 & $8.31\times$ \\ 
    1-30B & 48.90 & 3.15 & 1.61 & 60.19 & 6.89 & $8.74\times$ \\ 
    1-65B & 42.91 & 2.82 & 1.61 & 121.11 & 13.66 & $8.87\times$ \\ 
    2-7B & 46.95 & 3.03 & 1.61 & 12.31 & 1.56 & $7.89\times$ \\ 
    2-13B & 47.32 & 3.06 & 1.61 & 23.94 & 2.82 & $8.49\times$ \\ 
    2-70B & 47.93 & 3.11 & 1.61 & 145.49 & 16.16 & $9.00\times$ \\ 
    3-8B & 44.71 & 2.91 & 1.61 & 13.98 & 2.40 & $5.83\times$ \\ 
    \bottomrule
\end{tabular}
\caption{\textbf{Model-level pruning-quantization trade-off analysis.} 
We report the model-level pruning \textit{ratio} and quantization bit-width for remaining components (\textit{R-bit}) trade-off as well as an bits-per-weight (\textit{Bit}). For weight memory footprint (GB), we include an embedding layer and a normalization layer, which leads to a total size reduction (\textit{Size ↓}) of less than $10\times$ against FP16, even though we compress the weight matrix by the bits-per-weight of $1.61$.
}
\label{tab:tradeoff}
\end{table*}
}

\newcommand{\tabletradeofftwo}{
\begin{table*}[t]
\centering
\footnotesize
\setlength{\tabcolsep}{6pt}
\renewcommand{\arraystretch}{0.9}
\begin{tabular}{lccccccc}
    \toprule
    \multirow{2}{*}{\textbf{Module}} & \multirow{2}{*}{\specialcell{\textbf{SVD}\\\textbf{entropy}}} & \multicolumn{3}{c}{\textbf{Trade-off}} & \multicolumn{3}{c}{\textbf{Weight memory (MB)}} \\
    \cmidrule(lr){3-5}
    \cmidrule(lr){6-8}
    & & \textit{Ratio} & \textit{R-bit} & \textit{Bit} & \textit{FP16} & \textit{Ours} & \textit{Size ↓} \\ 
    \midrule
    Q & 6.64 & 91.80 & 1.89 & 0.08 & 32.00 & 0.32 & $103.37\times$ \\ 
    K & 6.53 & 86.62 & 1.78 & 0.12 & 32.00 & 0.48 & $67.15\times$ \\ 
    V & 7.84 & 45.61 & 3.34 & 0.91 & 32.00 & 3.63 & $8.81\times$ \\ 
    O & 7.85 & 62.26 & 3.52 & 0.66 & 32.00 & 2.66 & $12.04\times$ \\ 
    Up & 8.17 & 45.53 & 3.06 & 1.21 & 86.00 & 8.96 & $9.61\times$ \\ 
    Gate & 8.16 & 45.67 & 3.02 & 1.20 & 86.00 & 8.82 & $9.75\times$ \\ 
    Down & 8.22 & 38.53 & 4.23 & 1.90 & 86.00 & 13.99 & $6.15\times$ \\ 
    \midrule
    Avg. & 7.63 & 52.64 & 3.40 & 1.61 & 55.14 & 5.55 & $9.94\times$ \\
    \bottomrule
\end{tabular}
\caption{\textbf{Matrix-level pruning-quantization trade-off analysis.} 
We evaluate the matrix-level trade-off of pruning (ratio) and quantization for remaining components (r-bit) as well as an bits-per-weight (Bit). We use the LLaMA-2-7B model. 
}
\label{tab:tradeofftwo}
\end{table*}
}

\newcommand{\tableBiLevelAnalysis}{
\begin{table*}[t]
\centering
\footnotesize
\setlength{\tabcolsep}{6pt}
\renewcommand{\arraystretch}{0.9}
\begin{tabular}{ccccccc}
\toprule
\multirow{2}{*}{\textbf{Module}} & \multirow{2}{*}{\specialcell{\textbf{FP16}\\\textbf{(MB)}}} & \multirow{2}{*}{\specialcell{\textbf{SVD}\\\textbf{entropy}}} & \multicolumn{2}{c}{\specialcell{\textbf{Single-level}\\(\textit{Acc.: 48.77})}} & \multicolumn{2}{c}{\specialcell{\textbf{Bi-level}\\(\textit{Acc.: 49.06})}} \\
\cmidrule(lr){4-5}
\cmidrule(lr){6-7}
& &  
& \textit{Bit} & \textit{Mem} 
& \textit{Bit} & \textit{Mem} \\
\midrule
Q & 32.00 & 6.64 & 1.61 & 3.22 & 0.08 & 0.32 \\
K & 32.00 & 6.53 & 1.61 & 3.22 & 0.12 & 0.48 \\
V & 32.00 & 7.84 & 1.61 & 3.22 & 0.91 & 3.63 \\
O & 32.00 & 7.85 & 1.61 & 3.22 & 0.66 & 2.66 \\
Up & 86.00 & 8.17 & 1.61 & 8.65 & 1.21 & 8.96 \\
Gate & 86.00 & 8.16 & 1.61 & 8.65 & 1.20 & 8.82 \\
Down & 86.00 & 8.22 & 1.61 & 8.65 & 1.90 & 13.99 \\
\midrule
Avg. & 55.14 & 7.63 & 1.61 & 5.55 & 1.61\textsuperscript{\textdagger} & 5.55 \\
\bottomrule
\end{tabular}
\caption{\textbf{Bi-level differentiable bit-width analysis.} 
We compare matrix-level bits-per-weight (\textit{Bit}) and memory footprint (MB) between single-level (component-level) and the bi-level (component- and matrix-level) bit-width learning method. We employ the LLaMA-2-7B model. Our bi-level method optimizes the matrix-level bits-per-weight to reduce the errors. 1.61\textsuperscript{\textdagger} denotes the weighted average bit-width to account for the differences between the number of elements in the weight matrices.}
\label{tab:bilevel}
\end{table*}
}
\newcommand{\figconcept}{
\begin{wrapfigure}[10]{r}{0.55\columnwidth} 
    \vspace{-13pt}
    \centering
    
    \begin{subfigure}{0.6017928287\linewidth}
        \includegraphics[width=\linewidth]{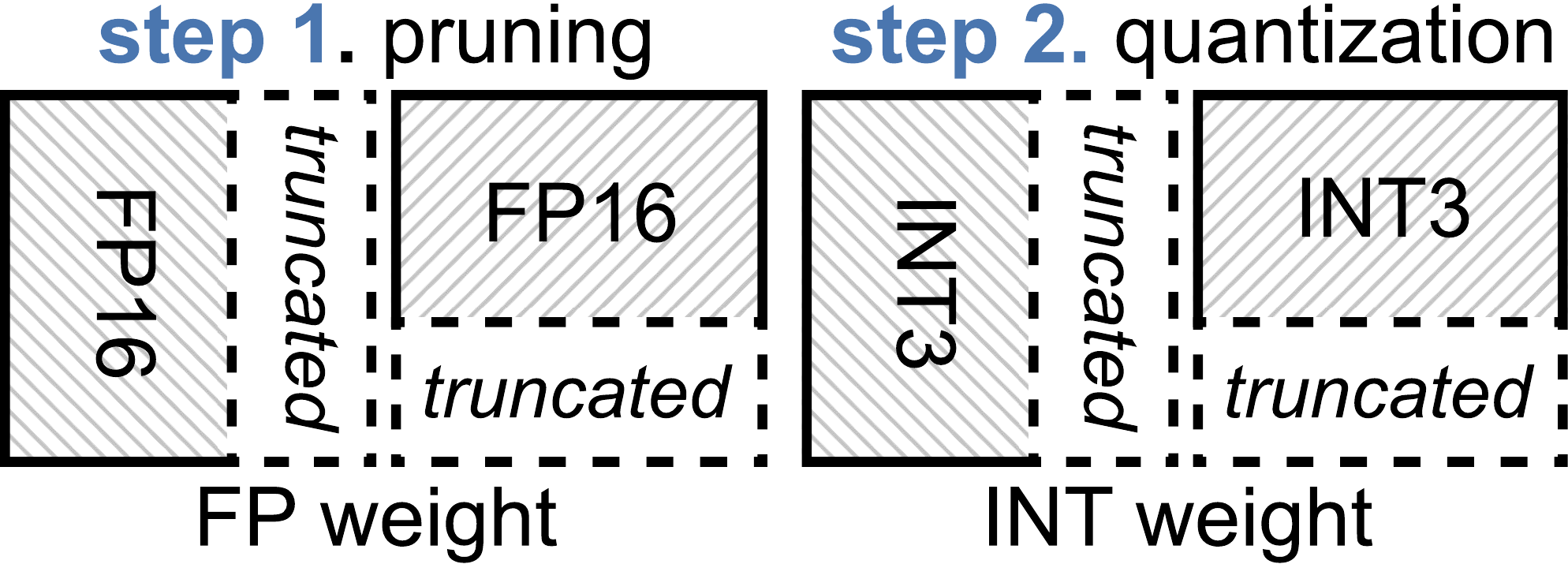}
        \caption{\textbf{Sequential} optimization}
        \label{fig:concept_1}
    \end{subfigure}
    \hfill
    \begin{subfigure}{0.3482071713\linewidth}
        \includegraphics[width=\linewidth]{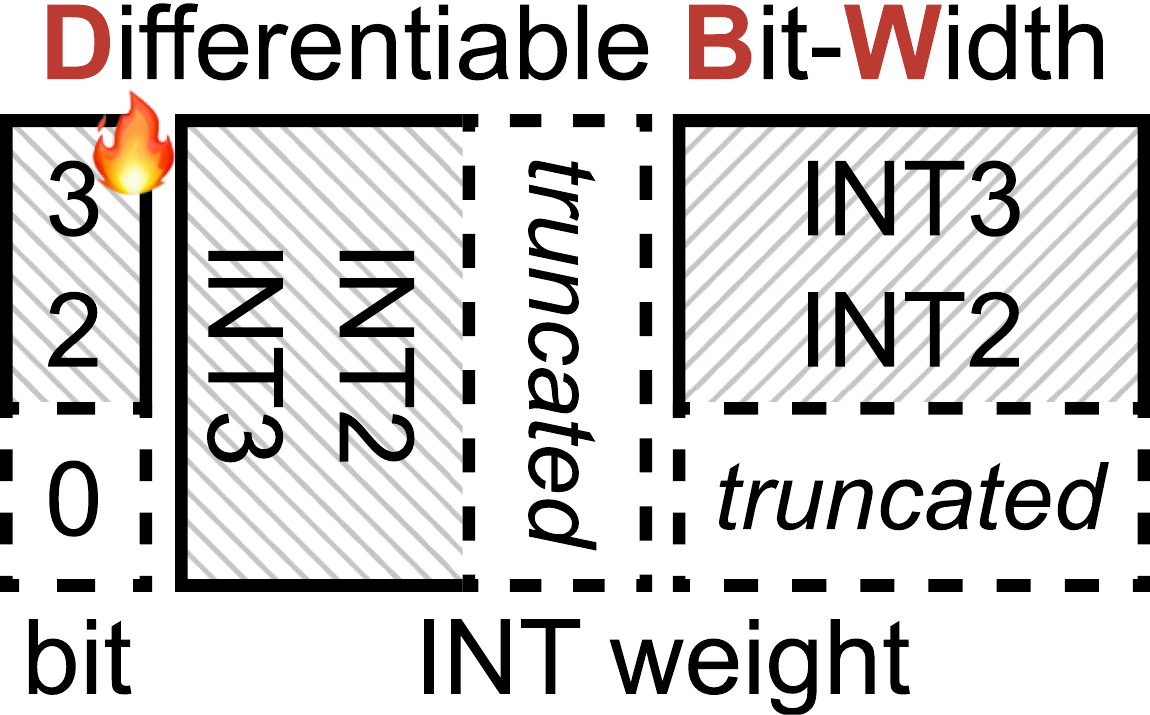}
        \caption{\textbf{Co}-optimization}
        \label{fig:concept_2}
    \end{subfigure}
    
    \caption{\textbf{Our compression concept illustration.} We co-optimize pruning and quantization in a single step by updating bit-widths in our differentiable method.
    }
    \label{fig:concept}
\end{wrapfigure}
}

\newcommand{\figframework}{
\begin{wrapfigure}[11]{r}{0.6\columnwidth}
    \vspace{-17pt}
    \centering
    \begin{subfigure}{0.4090019569\linewidth}
        \includegraphics[width=\linewidth]{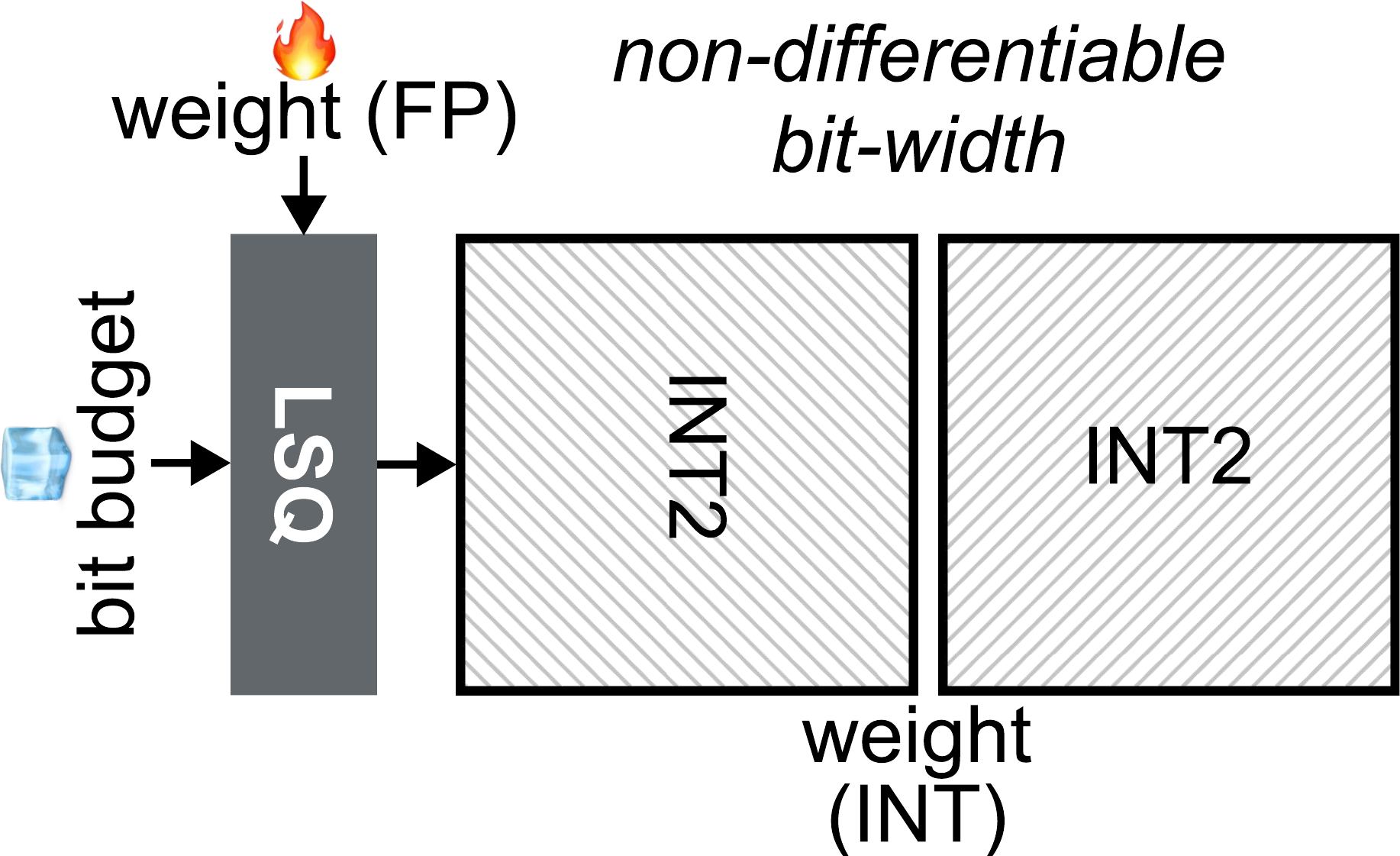}
        \caption{Non-differentiable bit}
        \label{fig:framework_1}
    \end{subfigure}
    \hfill
    \begin{subfigure}{0.5409980431\linewidth}
        \includegraphics[width=\linewidth]{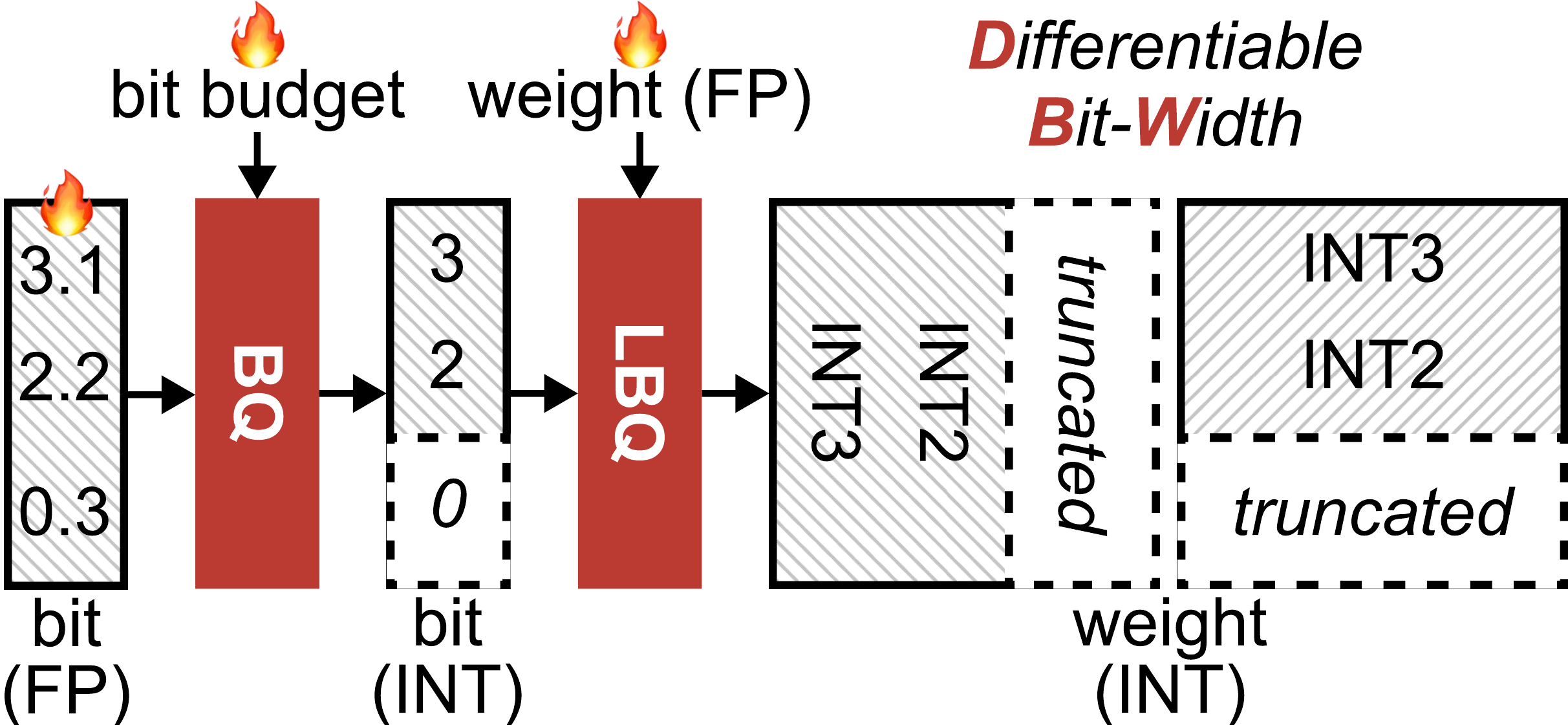}
        \caption{Our differentiable bit}
        \label{fig:framework_2}
    \end{subfigure}
    \caption{\textbf{Our differentiable method illustration.} (a) Non-differentiable method uses a standard quantizer~\citep{bhalgat2020lsq+} for uniform precision. (b) \ours~learns bit-width using the bit-width quantizer (BQ) and learned bit-width quantizer (LBQ). 
    }
    \label{fig:framework}
\end{wrapfigure}
}

\newcommand{\figmemory}{
\begin{figure}
    \centering
    \includegraphics[width=\linewidth]{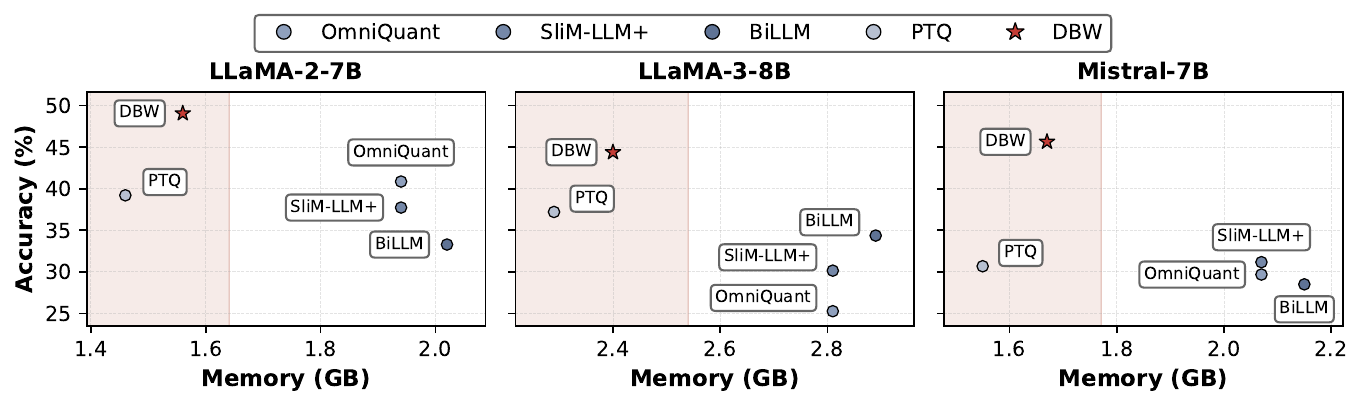}
    \caption{\textbf{Memory-accuracy trade-off evaluation.} We compare our DBW with representative ultra-low-bit baselines on LLaMA-2-7B, LLaMA-3-8B, and Mistral-7B. Each point shows the model memory footprint (GB) and the mean accuracy on the eight commonsense reasoning benchmarks. The shaded regions highlight the extreme memory-compression regimes (\textit{e.g.}, $7.5\times$, $5.5\times$, and $7.5\times$).
    }
    \label{fig:memory}
\end{figure}
}

\newcommand{\figlatency}{
\begin{wrapfigure}[10]{r}{0.48\columnwidth}
    \centering
    \vspace{-2em}
    \includegraphics[width=\linewidth]{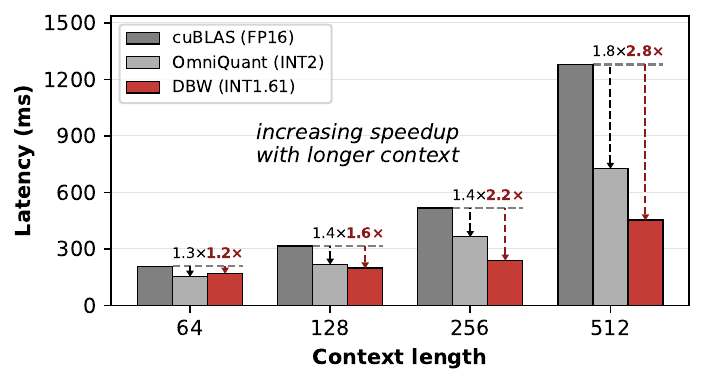}
    \vspace{-2em}
    
    \caption{
    \textbf{End-to-end latency evaluation.} 
    }
    \label{fig:latency}
\end{wrapfigure}
}

\newcommand{\figtradeoffcomponent}{
\begin{wrapfigure}[7]{r}{0.45\columnwidth} 
    \vspace{-2em}
    \centering
    \includegraphics[width=\linewidth]{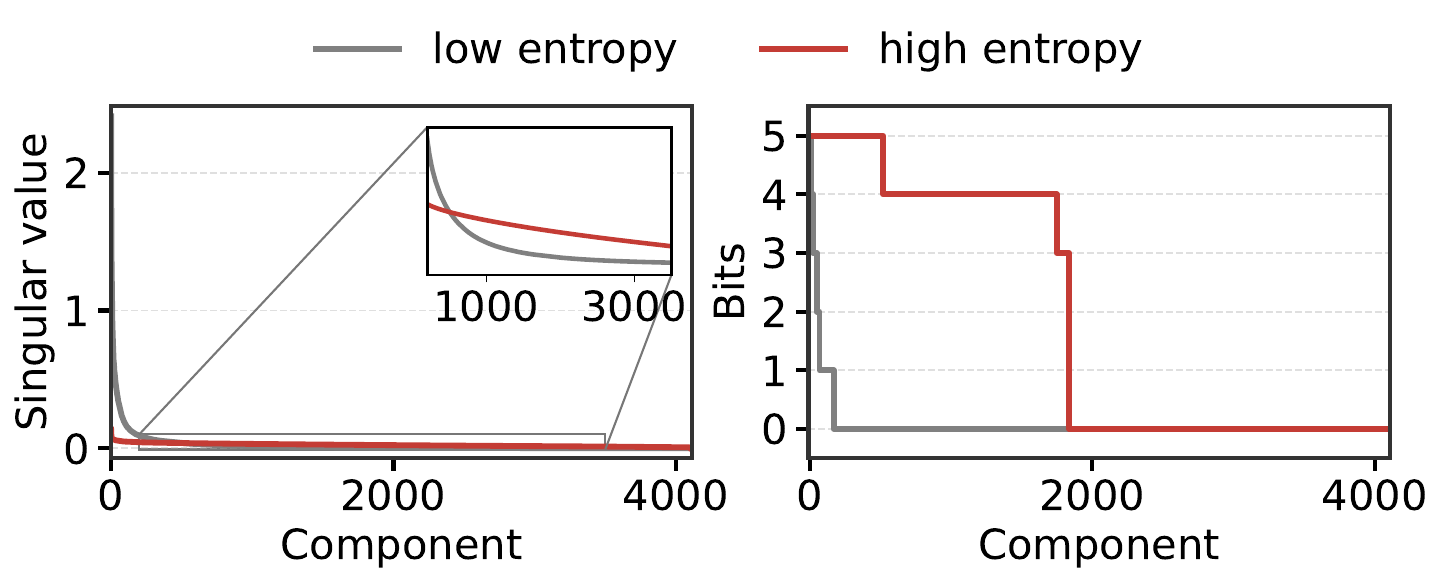}
    \vspace{-1.8em}
    \caption{\textbf{Learned bit-width analysis.} 
    }
    \label{fig:tradeoffcomponent}
\end{wrapfigure}
}

\newcommand{\figoptimaltradeoff}{
\begin{figure*}[h]    
    \centering

    \begin{subfigure}{0.25\linewidth}
        \includegraphics[width=\linewidth]{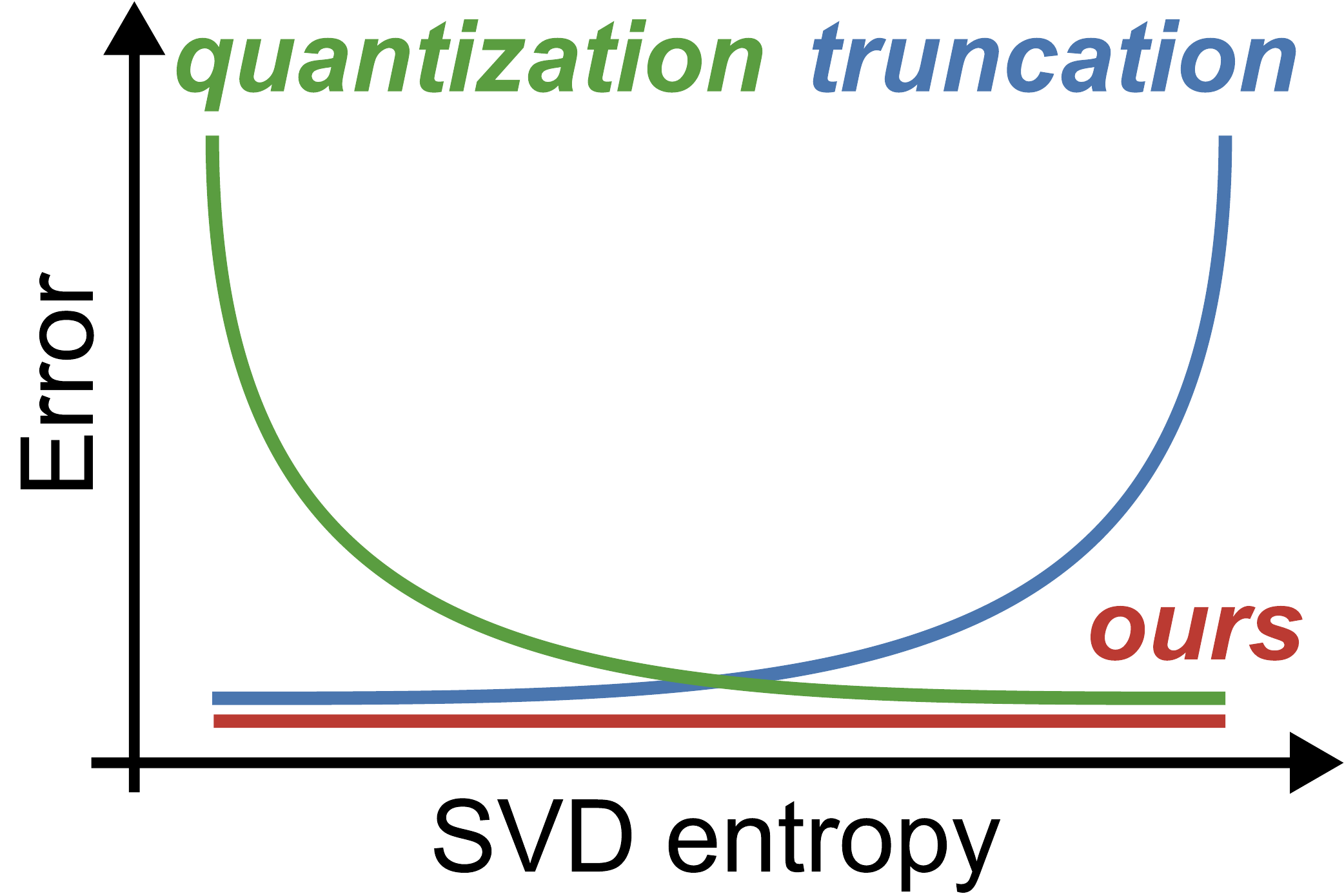}
        \caption{Singular value}
        \label{fig:optimaltradeoff_1}
    \end{subfigure}
    \begin{subfigure}{0.25\linewidth}
        \includegraphics[width=\linewidth]{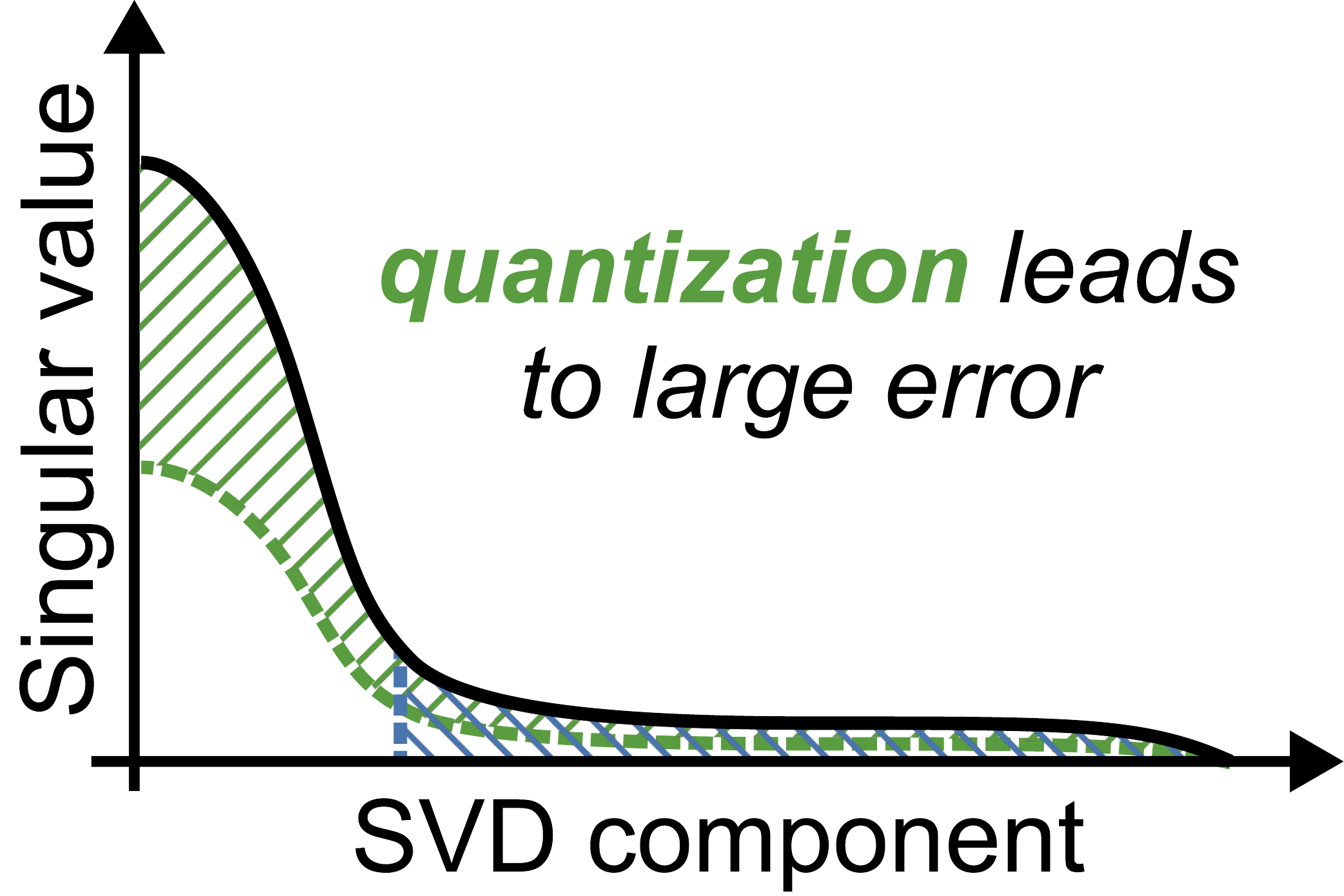}
        \caption{Low entropy}
        \label{fig:optimaltradeoff_2}
    \end{subfigure}
    \begin{subfigure}{0.25\linewidth}
        \includegraphics[width=\linewidth]{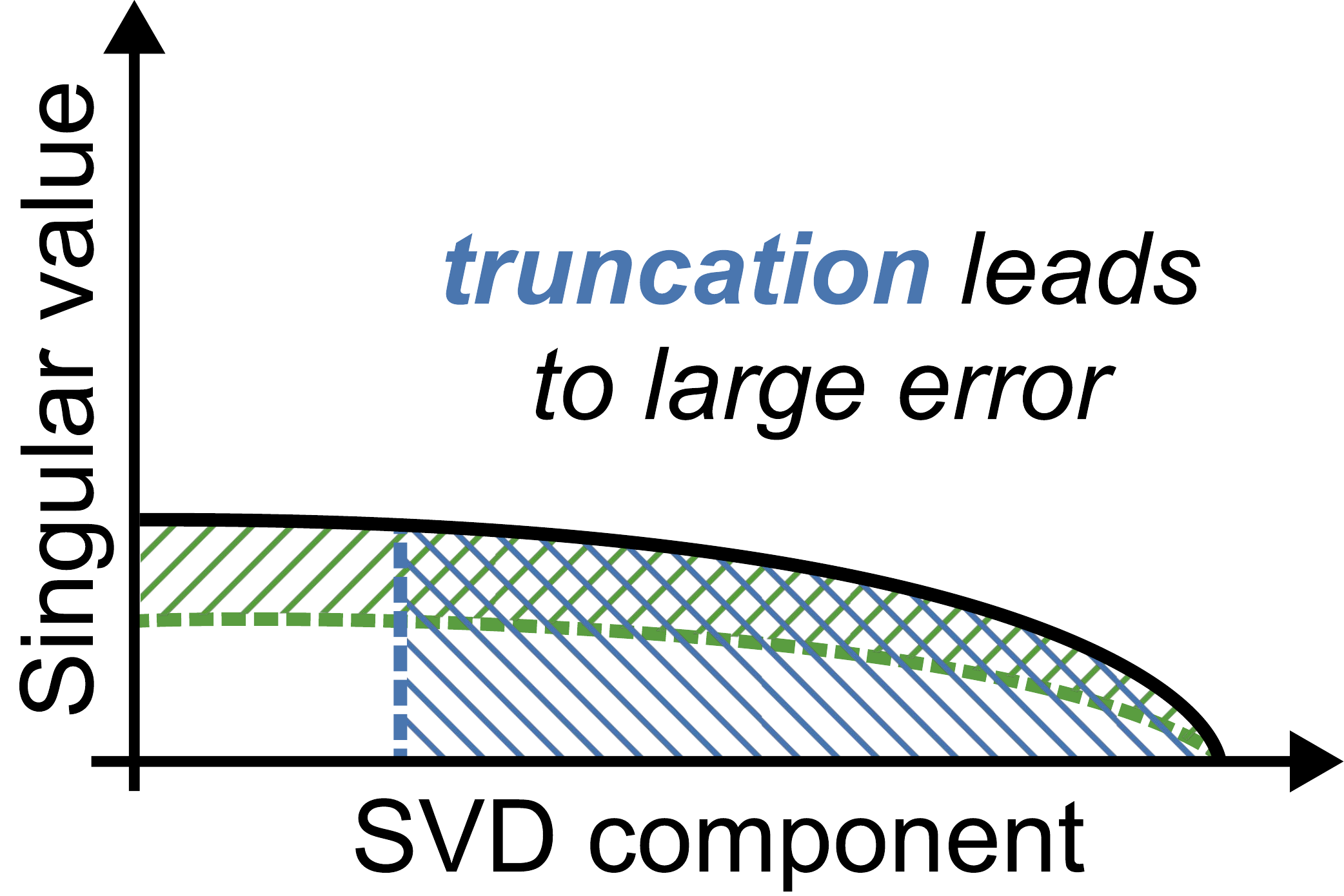}
        \caption{High entropy}
        \label{fig:optimaltradeoff_3}
    \end{subfigure} 
    
    \caption{\textbf{Optimal trade-off concept illustration.} 
    (a) We depict the reconstruction error of truncation and quantization as a function of singular-value entropy. (b) For the low SVD entropy, the quantization leads to a high error rate due to assigning uniform precision across components. (c) When the SVD entropy is high, the truncation error becomes severe due to removing a considerable information that is evenly distributed among components.
    }
    \label{fig:optimaltradeoff}
\end{figure*}
}

\newcommand{\figsingularvalue}{
\begin{figure}[h]    
    \centering
    \includegraphics[width=0.5\linewidth]{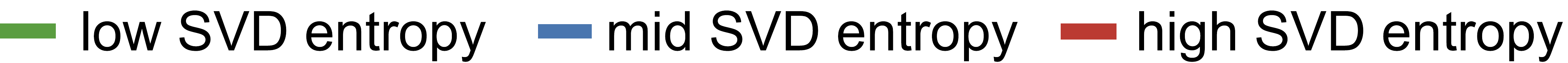}
    
    \begin{subfigure}{0.25\linewidth}
        \includegraphics[width=\linewidth]{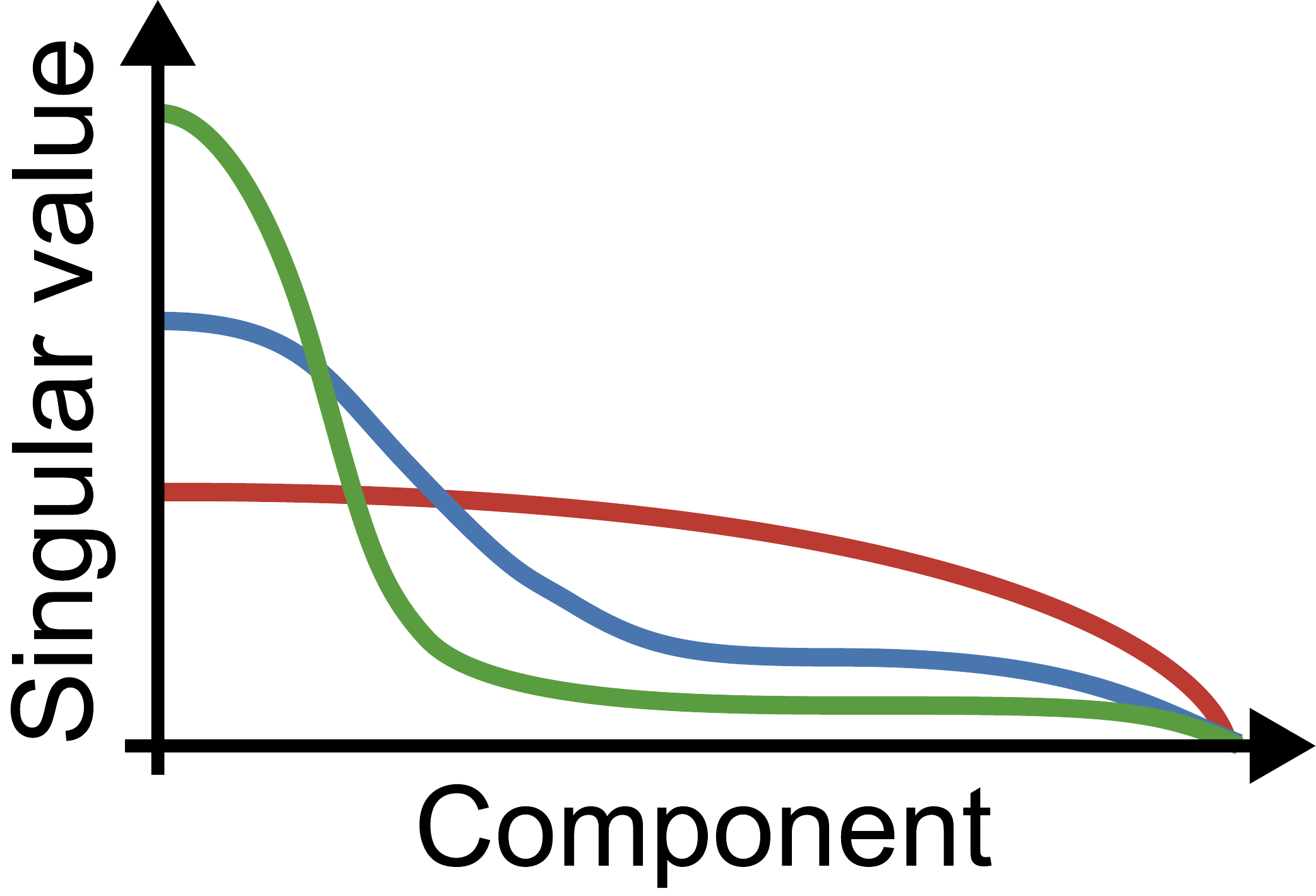}
        \caption{Singular value}
        \label{fig:singularvalue_1}
    \end{subfigure}
    \begin{subfigure}{0.25\linewidth}
        \includegraphics[width=\linewidth]{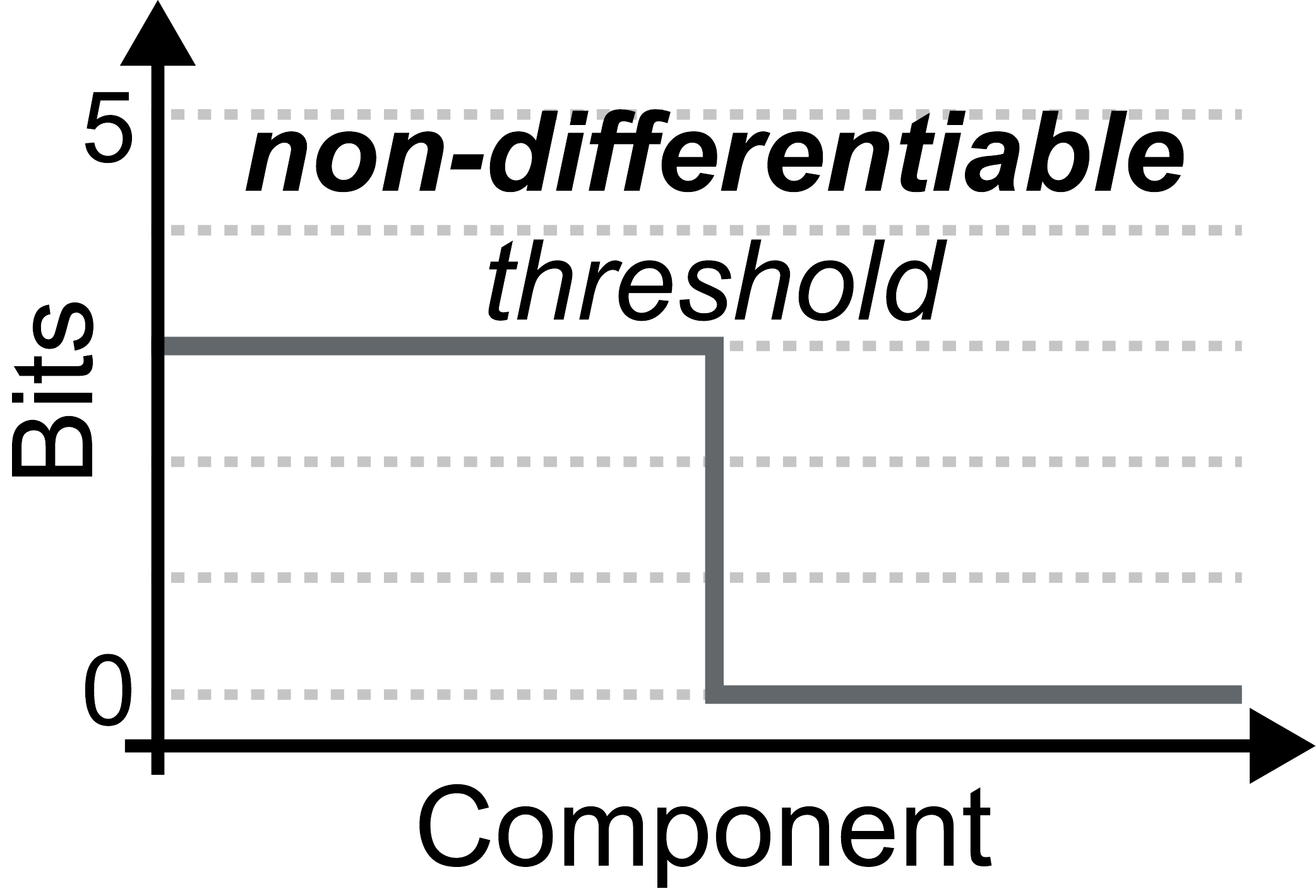}
        \caption{Static}
        \label{fig:singularvalue_2}
    \end{subfigure}
    \begin{subfigure}{0.25\linewidth}
        \includegraphics[width=\linewidth]{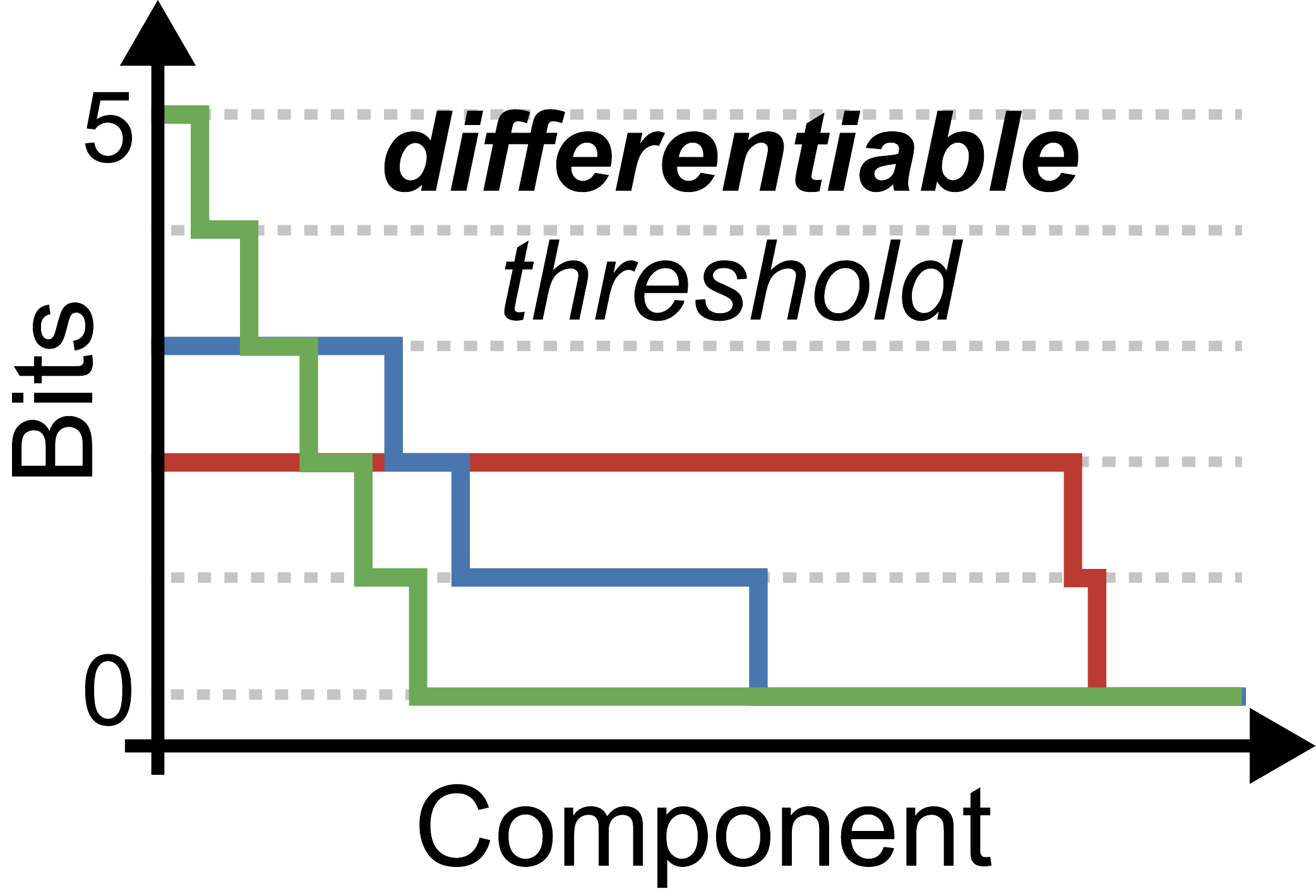}
        \caption{Adaptive}
        \label{fig:singularvalue_3}
    \end{subfigure} 
    
    \caption{\textbf{Adaptive mixed-bit allocation illustration.} 
    While the singular value distribution dictates the optimal bit-width, conventional static methods rely on non-differentiable thresholds. In contrast, our adaptive approach assigns differentiable thresholds based on the singular value distribution, naturally facilitating both pruning (bit=0) and quantization (bit>0) in a unified manner.
    }
    \label{fig:singularvalue}
\end{figure}
}

\newcommand{\figlbq}{
\begin{figure*}[h]
    \centering
    \includegraphics[width=0.9\linewidth]{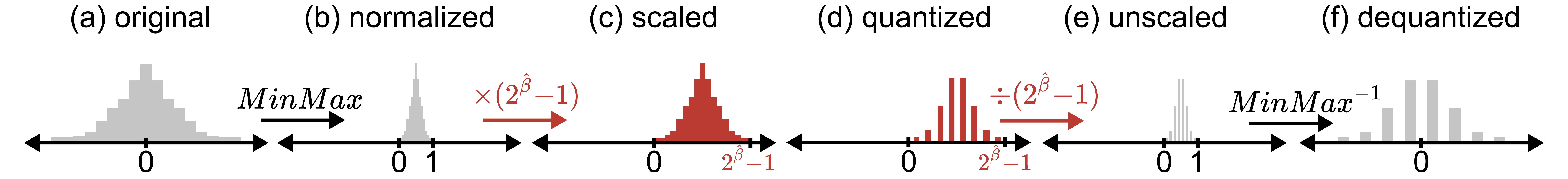}
    \caption{\textbf{Our learned bit quantizer illustration.} We use the learned bit quantizer to allow the backpropagation for the bit-width parameters.}
    \label{fig:lbq}
\end{figure*}
}

\newcommand{\figlbqgradient}{
\begin{figure*}[h]
    \centering
    \includegraphics[width=0.4\linewidth]{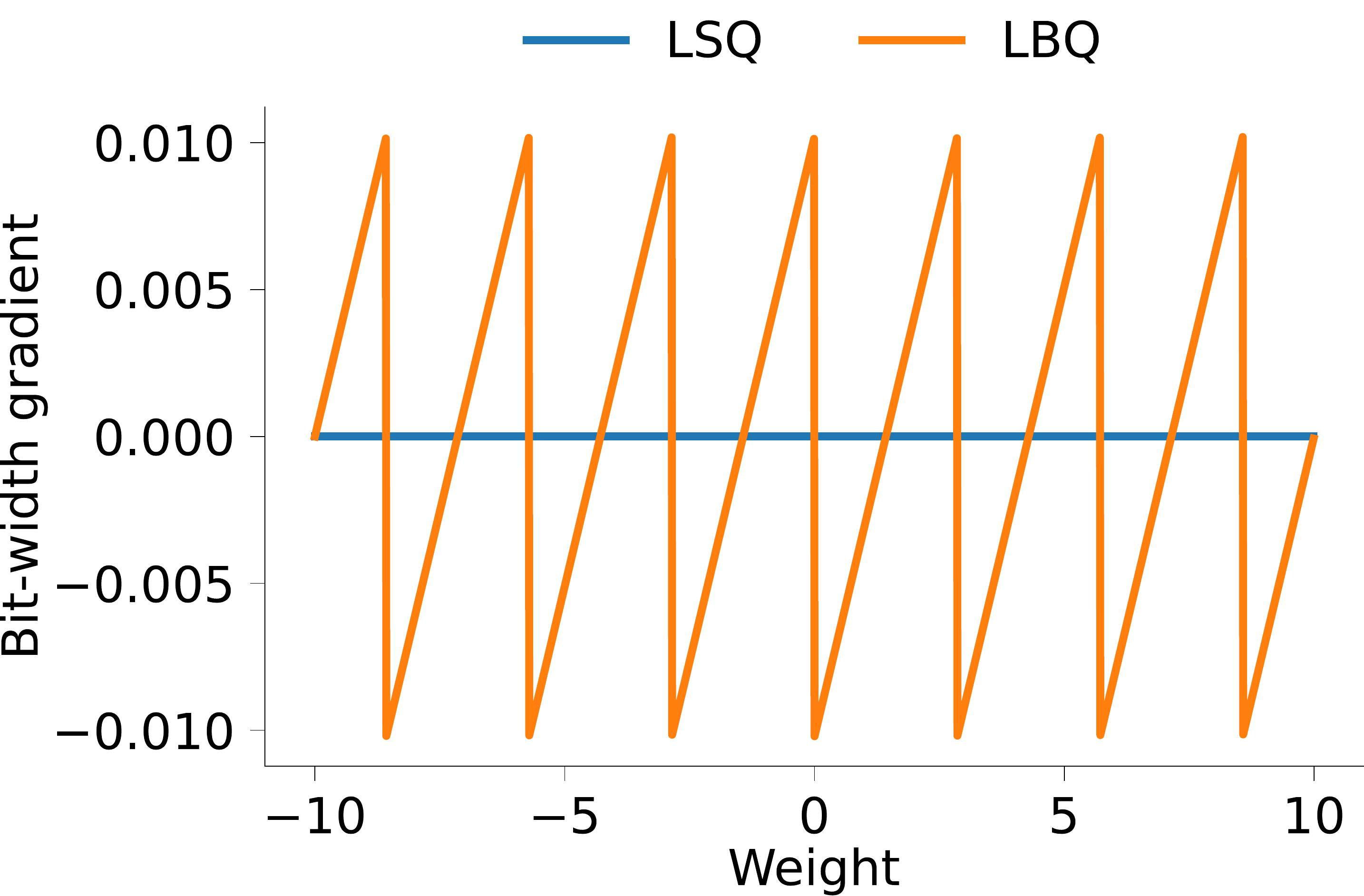}
    \caption{\textbf{Gradient analysis.} We analyze the gradient of bit-width parameters as a function of the weight values. Compared to the commonly used LSQ method, the proposed method can provide the gradient of rounding error to the bit-width parameters.}
    \label{fig:lbq_gradient}
\end{figure*}
}

\newcommand{\figbittrajectory}{
\begin{figure*}[h]
    \centering
    \includegraphics[width=0.5\linewidth]{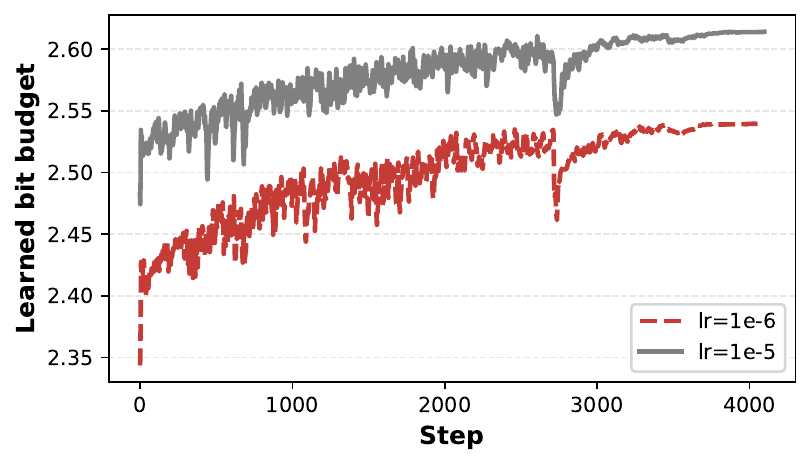}
    \caption{\textbf{Trajectory of learned bit budget.} We visualize the learned bit budgets during calibration on the first MLP layer of LLaMA-2-7B under two matrix-level bit learning rates. Across both settings, the learned bit-budgets remain far from zero and gradually converge to a stable range, indicating that the matrix-level budget does not collapse during optimization. This supports our observation that the reconstruction loss acts as an implicit lower-bound regularizer for the bit-budget parameters.}
    \label{fig:trajectory}
\end{figure*}
}

\newcommand{\figdequant}{
\begin{figure}[h]
    \centering
    \includegraphics[width=0.55\linewidth]{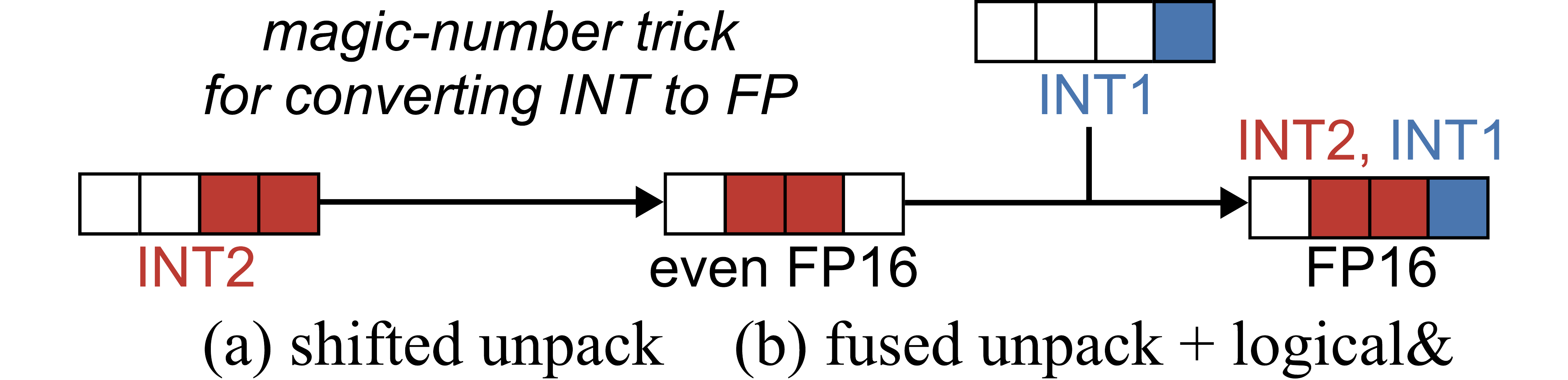}
    \caption{\textbf{Our on-the-fly dequantization kernel.} Our on-the-fly dequantization kernel converts a pair of separated integer values (\textit{e.g.}, INT2 and INT1 in this case) into FP16 format by efficiently fusing them via (a) shift and (b) logical $\&$ operator.}
    \label{fig:dequant}
\end{figure}
}

\newcommand{\figtradeoffblock}{
\begin{figure*}[t]
    \centering
    \hfill\hfill
    \begin{subfigure}{0.23\textwidth}
        \includegraphics[width=\linewidth]{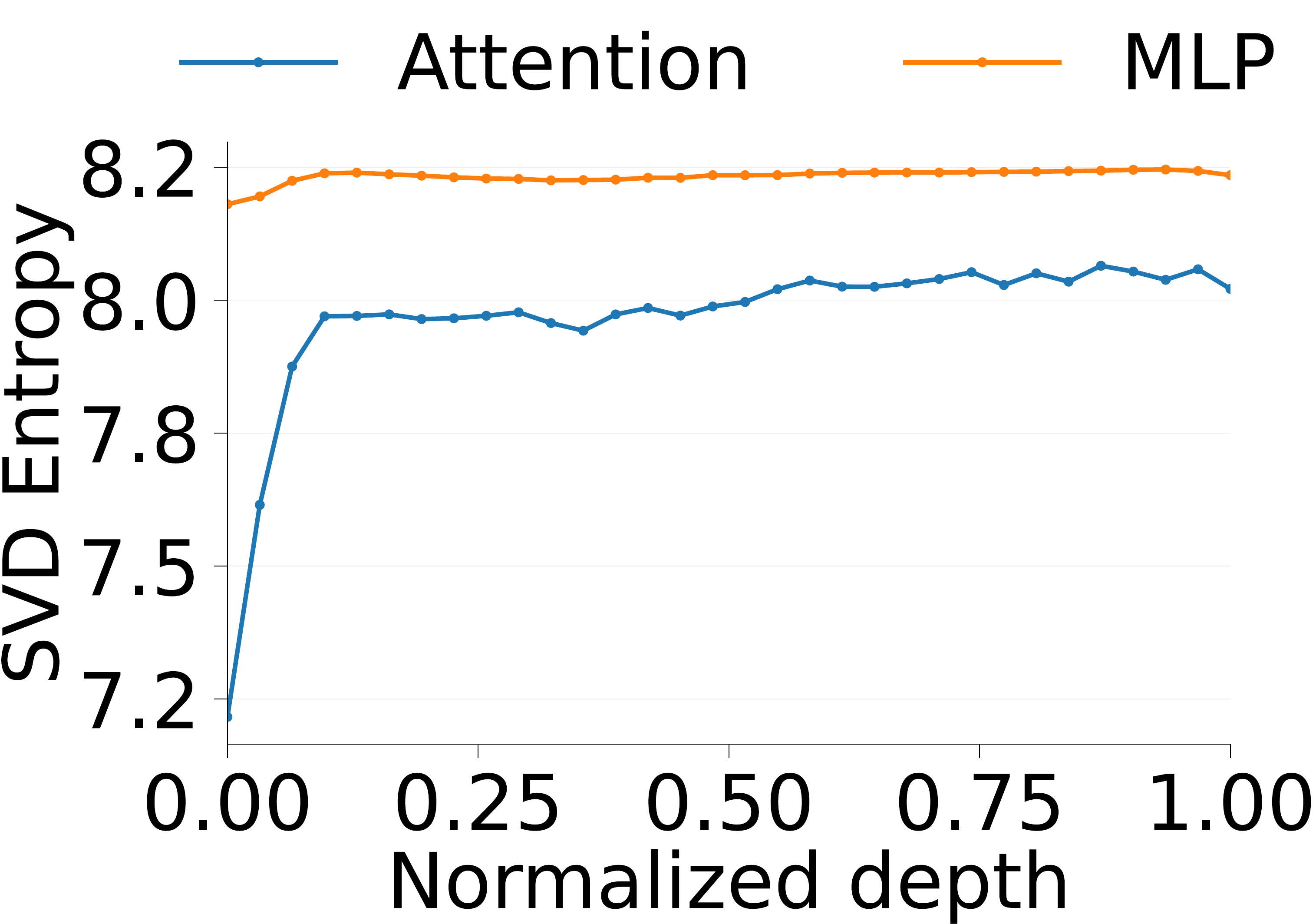}
        \caption{SVD entropy}
        \label{fig:tradeoffblock_1}
    \end{subfigure}
    \hfill
    \begin{subfigure}{0.23\linewidth}
        \includegraphics[width=\linewidth]{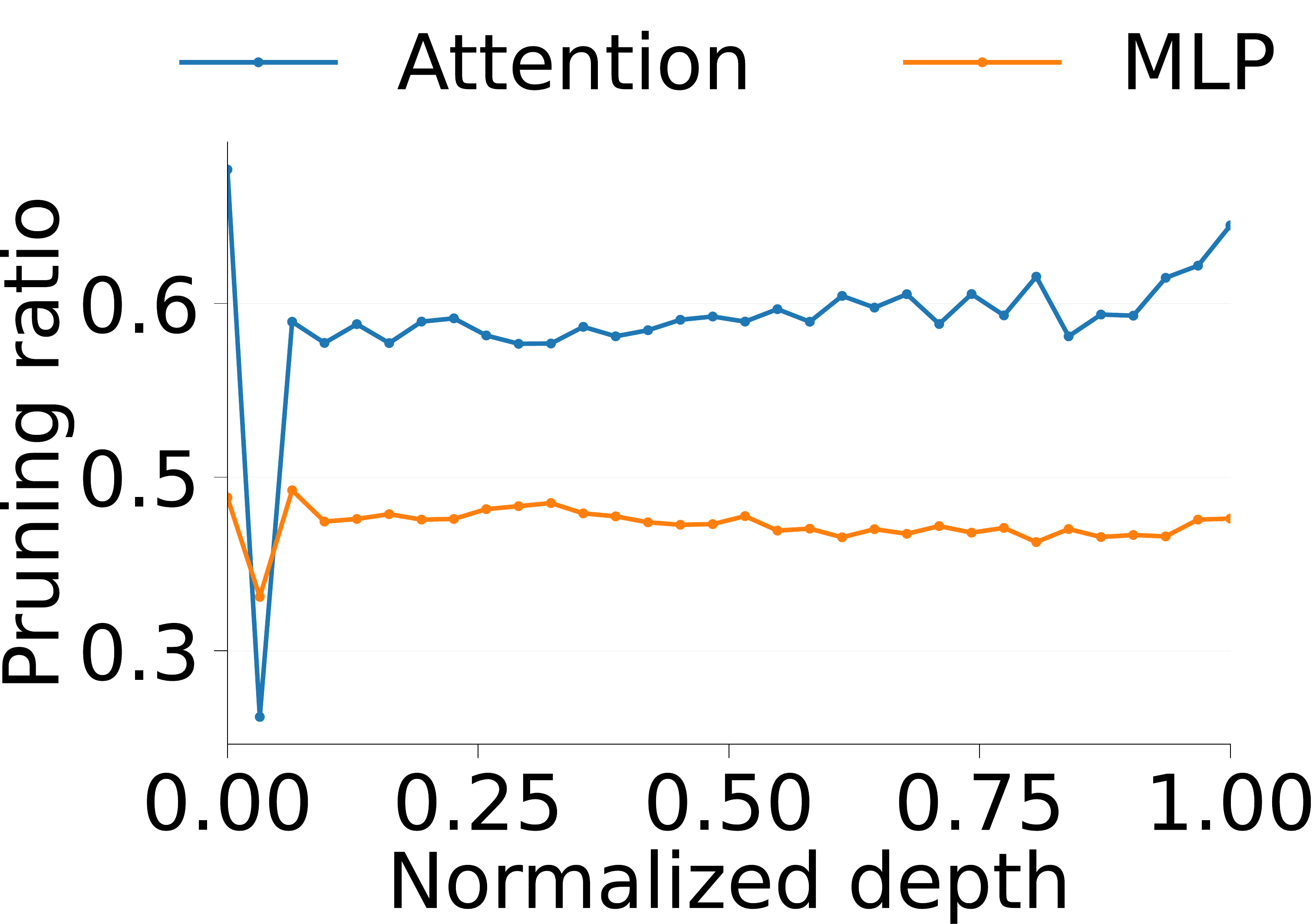}
        \caption{Pruning ratio}
        \label{fig:tradeoffblock_2}
    \end{subfigure}
    \hfill
    \begin{subfigure}{0.23\linewidth}
        \includegraphics[width=\linewidth]{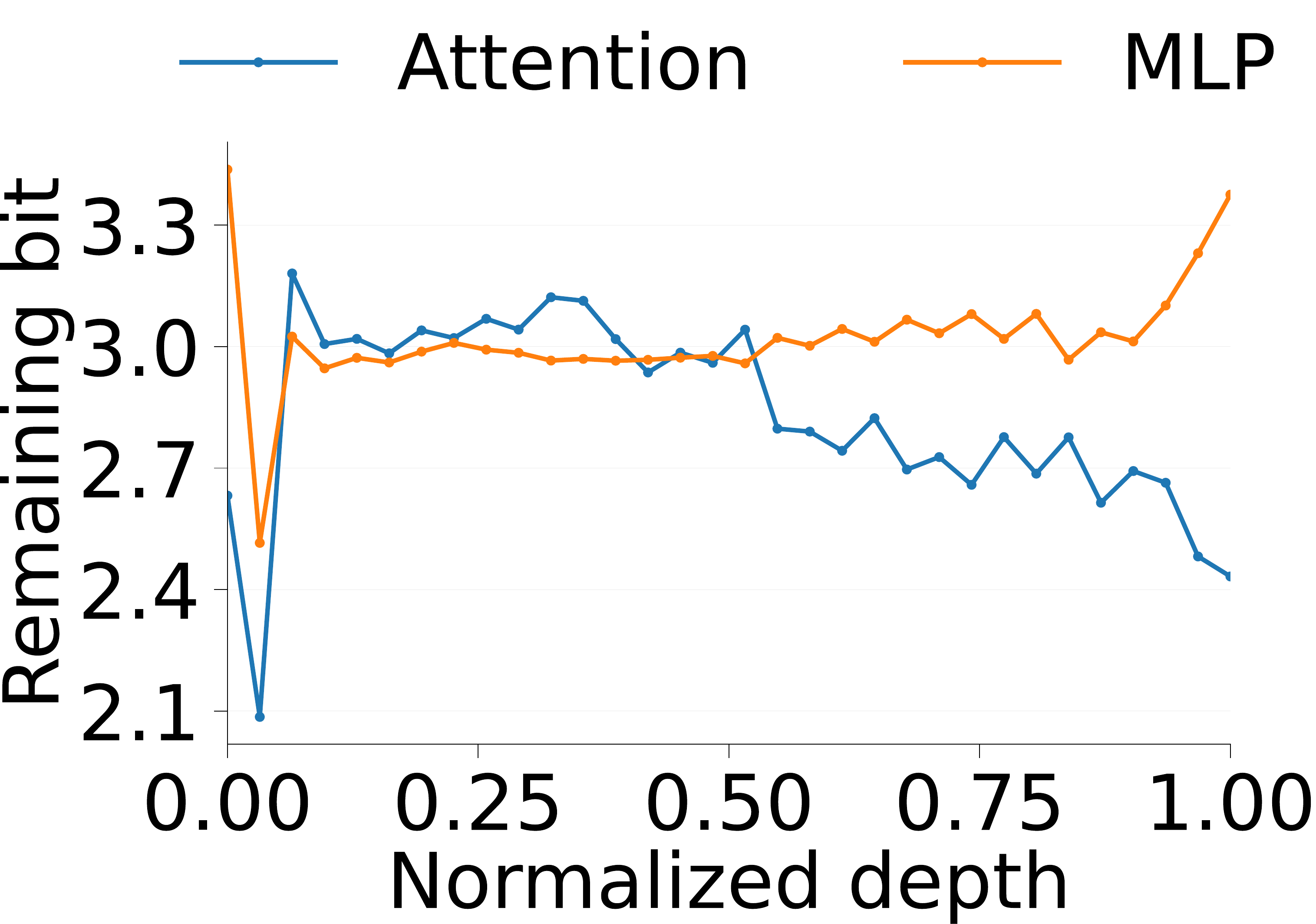}
        \caption{Remaining bit}
        \label{fig:tradeoffblock_3}
    \end{subfigure}
    \hfill
    \begin{subfigure}{0.23\linewidth}
        \includegraphics[width=\linewidth]{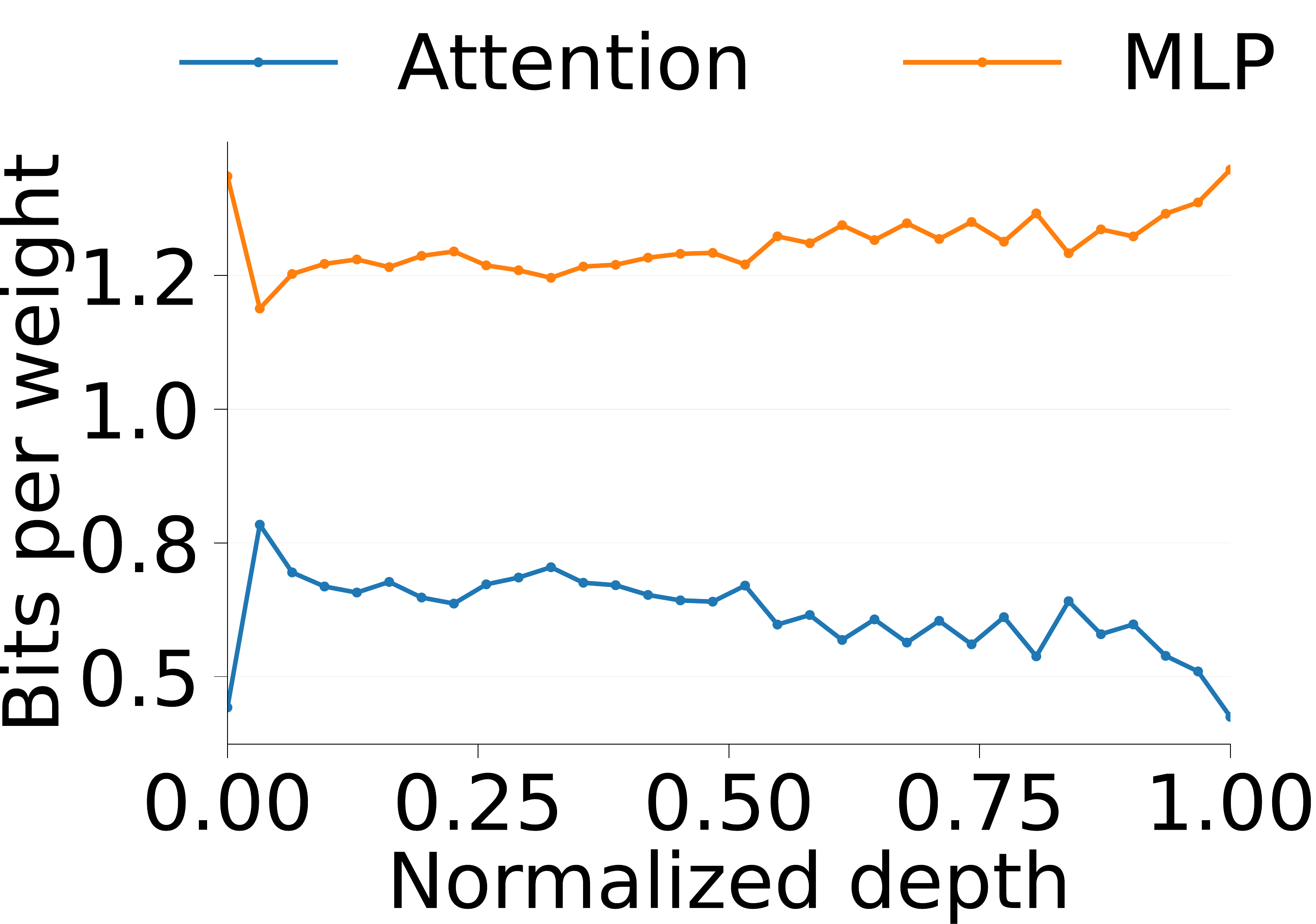}
        \caption{Bits per weight}
        \label{fig:tradeoffblock_4}
    \end{subfigure} 
    \hfill\hfill\hfill
    
    \caption{\textbf{Block-level truncation-quantization trade-off analysis.} We report the block-level truncation and quantization trade-off. We use the LLaMA-2-7B model.}
    \label{fig:tradeoffblock}
\end{figure*}
}

\newcommand{\figsvdentropy}{
\begin{figure}[t]
    \centering
    \begin{subfigure}{0.3\linewidth}
        \includegraphics[width=\linewidth]{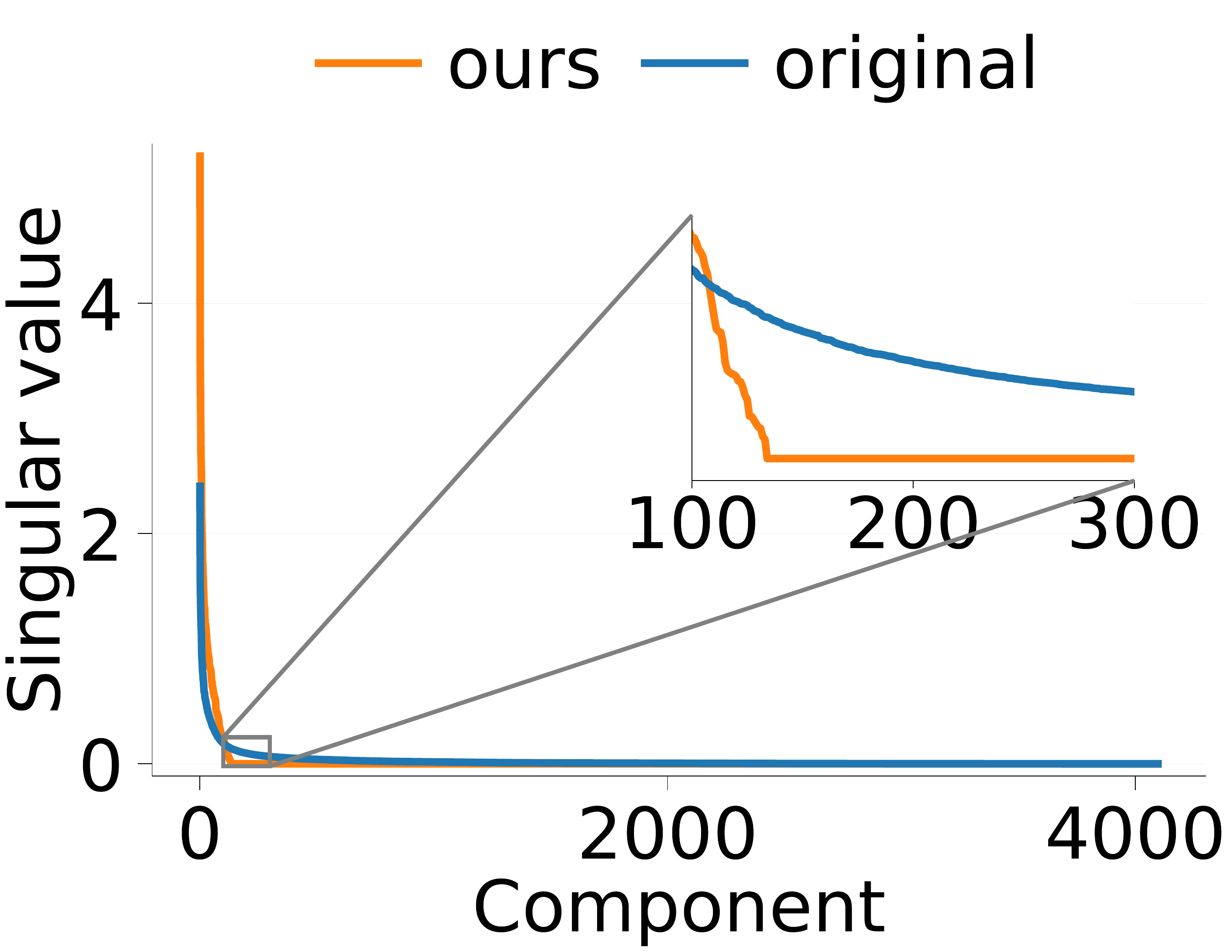}
        \caption{Low SVD entropy}
        \label{fig:svdentropy_1}
    \end{subfigure}
    \begin{subfigure}{0.3\linewidth}
        \includegraphics[width=\linewidth]{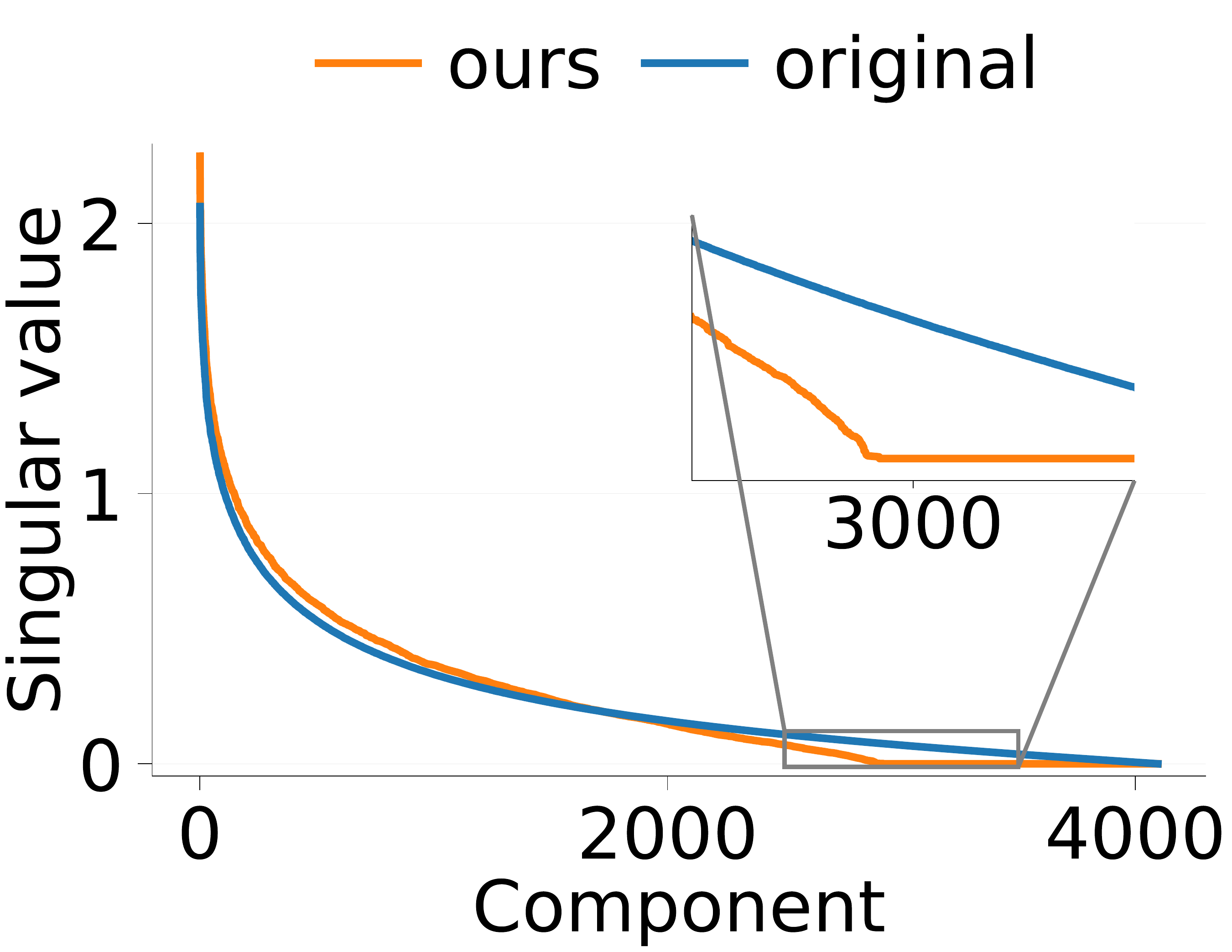}
        \caption{High SVD entropy}
        \label{fig:svdentropy_2}
    \end{subfigure}
    \caption{\textbf{Low SVD entropy due to our loss.} We show that our entropy minimization loss effectively reduces the SVD entropy compared to the baseline using LLaMA-2-7B model.}
    \label{fig:svdentropy}
\end{figure}
}
\newcommand{\algfiggemm}{
\begin{figure*}[t]
\centering

\begin{minipage}[t]{0.49\textwidth}
\vspace{0pt}
\rule{\linewidth}{0.8pt}\par
\vspace{-1pt}
\textbf{Algorithm 1} Pseudocode of our learning method.\par
\label{alg:svq}
\vspace{-6pt}
\rule{\linewidth}{0.4pt}\par
\vspace{-3pt}

\begin{algorithmic}[1]
\footnotesize
\Require block parameters $\theta=\{\theta^i\}_{i=1}^{M}$,
block function $f$, inputs $\mathcal{X}$,
learning rate $\gamma$

\State $\{\Theta^i\}_{i=1}^{M}
=
\{\mathrm{SVD}(\theta^i)\}_{i=1}^{M}$
\Comment{$\Theta^i=\{\mathbf{U}^i,\Sigma^i,\mathbf{V}^i\}$}

\State Initialize bit-width parameters $\{\alpha^i,\beta^i\}_{i=1}^{M}$

\For{number of training iterations}
\State $l_\text{ent} \gets 0$
\For{$m=1$ to $M$} 
    \State $\hat{\alpha}^{m} \gets \mathrm{BA}(\alpha^{m})$
    \Comment{matrix-bit of Eq.~(\textcolor{BrickRed}{3})}
    \For{$c=1$ to $C^{m}$}
        \State $\hat{\beta}_{c}^{m} \gets \mathrm{BQ}(\beta_{c}^{m}, \hat{\alpha}^{m})$
        \Comment{comp.-bit of Eq.~(\textcolor{BrickRed}{1})}
        \State $\hat{\mathbf{U}}_{:,c}^{m} \gets \mathrm{LBQ}(\mathbf{U}_{:,c}^{m}, \hat{\beta}_{c}^{m})$
        \State $\hat{\mathbf{V}}_{:,c}^{m} \gets \mathrm{LBQ}(\mathbf{V}_{:,c}^{m}, \hat{\beta}_{c}^{m})$
    \EndFor
    \State $l_\text{ent} \gets l_\text{ent} + L_\text{ent}(\Sigma^{m})$
    \Comment{entropy loss of Eq.~(\textcolor{BrickRed}{4})}
    \State $\hat{\Theta}^{m} \gets \hat{\mathbf{U}}^{m} {\Sigma}^{m} \hat{\mathbf{V}}^{m\top}$
\EndFor

\State $l_{\mathrm{rec}} \gets 
L_\text{rec}(\mathcal{X},f,\{\hat{\Theta}^{i},\theta^{i}\}_{i=1}^{M})$
\Comment{mse of Eq.~(\textcolor{BrickRed}{2})}

\State $\phi \gets \phi - \gamma \nabla_{\phi} (l_\text{rec}+l_\text{ent}) 
\;\; (\phi \in \{\Theta,\alpha,\beta\})$ 
\EndFor

\end{algorithmic}
\vspace{-6pt}
\rule{\linewidth}{0.8pt}

\end{minipage}
\hfill
\begin{minipage}[t]{0.49\textwidth}
\vspace{0pt}
\centering

\begin{subfigure}{0.85\linewidth}
    \includegraphics[width=\linewidth]{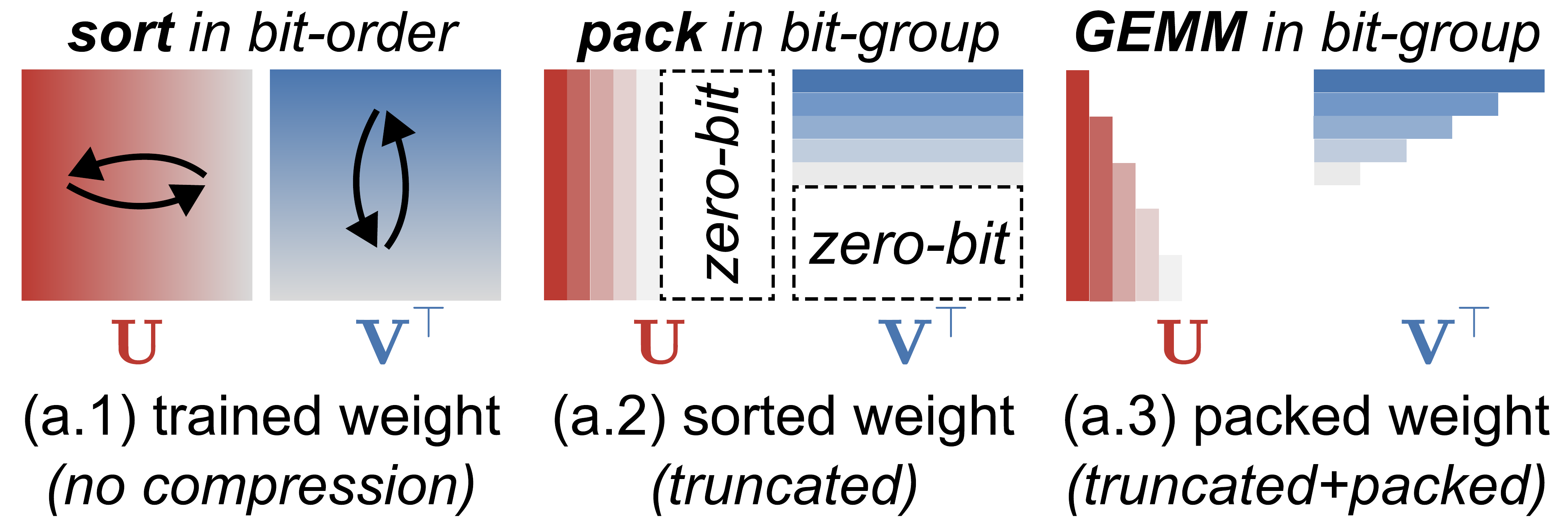}
    \caption{Bit-group-wise weight packing}
    \label{fig:gemm_1}
\end{subfigure}

\begin{subfigure}{0.85\linewidth}
    \includegraphics[width=\linewidth]{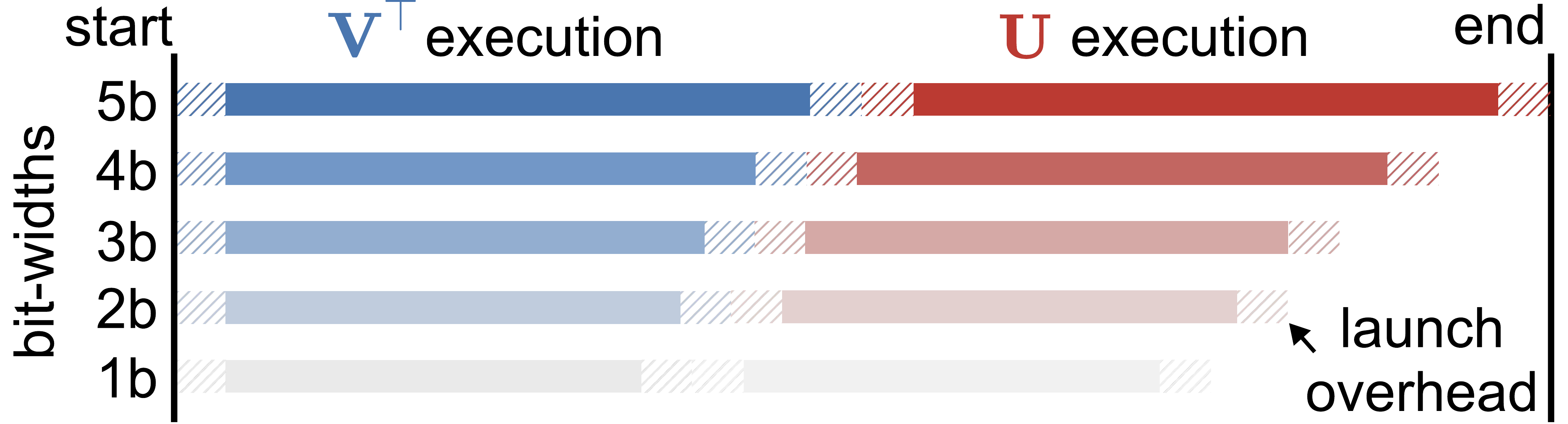}
    \caption{Parallel GEMM execution}
    \label{fig:gemm_2}
\end{subfigure}

\caption{\textbf{Our mixed-precision GEMM illustration.} 
Ours (a) compress the weight memory by bit-group-wise packing and 
(b) use parallel execution of bit-group GEMMs to maximize GPU.}
\label{fig:gemm}

\end{minipage}

\end{figure*}
}

\title{Differentiable Bit-Widths: Co-optimizing Pruning and Quantization via SVD for Ultra-Efficient LLM Compression}

\author{
  Hankyul Kang\\
  Ajou University\\
  \texttt{hankyulkang@ajou.ac.kr}
  \And
  Jongbin Ryu\thanks{Corresponding author.}\\
  Ajou University\\
  \texttt{jongbinryu@ajou.ac.kr}
}

\begin{document}

\maketitle

\begin{abstract}
SVD-based pruning and quantization have recently emerged as a promising strategy for the ultra-efficient compression of large language models. In these methods, compression is performed in two stages: components are first truncated, and the remaining ones are subsequently quantized. Although this decoupled pipeline benefits from both pruning and quantization, it requires separate optimization for each stage and fails to fully exploit their balance, which can lead to suboptimal performance under aggressive compression.
To address this limitation, we propose a new LLM compression method that co-optimizes pruning and quantization in a unified framework. Our key idea is a differentiable method for learning component-wise bit-widths, allowing less important components to be assigned 0-bit precision and pruned away. Notably, our method performs favorably against two-stage baselines, even when subjected to extreme quantization settings ($1.61$ bits) designed for ultra-efficiency. 
Code: \url{https://github.com/MMAI-Laboratory/DBW}.
\end{abstract}

\section{Introduction}
\label{sec:intro}
Ultra-efficient compression serves as a key enabler for the seamless deployment of large language models (LLMs) in real-world environments~\citep{zhu2024survey}. 
Given that LLMs should operate on edge devices using limited hardware resources, it is imperative that compression methods significantly reduce memory usage and latency to accommodate \han{such constraints}.
For example, to execute LLaMA-7B~\citep{touvron2023llama} on a consumer laptop GPU with 2GB of VRAM, we need to compress its parameters by approximately an order of magnitude.
This lowers the \han{memory footprint} from 12.31GB\footnote{We measure memory usage under FP16 computation settings.} to 1.56GB, facilitating its deployment on such a laptop.
However, it is widely recognized that reducing the parameters by more than $3\times$--$4\times$ will drastically degrade the performance of LLMs~\citep{ouyang2025low}.

Recently, model pruning and quantization using singular value decomposition (SVD) have emerged as a promising approach to ultra-efficient LLM compression~\citep{yuan2023asvd,wang2024svd,wang2025qsvd}.
They have shown that model parameters can be effectively pruned by truncating the low singular-value components from the decomposed matrix.
Additionally, the model can be further compressed by applying a quantization method to the non-truncated components.
Although this two-stage approach enjoys the benefits of both pruning and quantization, it has the limitation that it operates as a decoupled process, which requires separate optimization for each stage.
Obviously, the pruning ratio and quantization bit-width are closely related factors that play a critical role in optimizing LLM compression. A higher pruning ratio allows non-truncated components to be quantized at higher bit-widths, while a lower pruning ratio requires lower bit-widths. 
However, two-stage approaches that treat pruning and quantization as decoupled sequential steps often fail to find the exact balance, resulting in suboptimal trade-offs. 
This limitation is consistent with a widely recognized principle~\citep{lecun2002gradient,bahdanau2014neural} in machine learning: jointly optimizing multiple constraints typically yields better performance compared to isolated  sequential approaches.
In a sequential pipeline, the quantization stage cannot recover the representational capacity irreversibly lost during the pruning stage, causing compound errors. 
On the other hand, co-optimization facilitates the efficient exploration of the combined parameter space of a model, helping it to adeptly negotiate the complex dependencies between sparsity and precision.
Therefore, it is key to optimize pruning and quantization simultaneously to maximize their benefits for achieving ultra-efficient compression of LLMs.
Motivated by this, we propose a co-optimized compression method that tightly couples pruning and quantization. 
Specifically, the proposed method applies an adaptive bit-width strategy to SVD-decomposed matrices. 
By dynamically allocating mixed bit-widths to components, the 0-bit \han{precision} induces rank truncation, allowing us to seamlessly accomplish both pruning and quantization in a \han{unified step (see~\cref{fig:concept})}.

\figconcept

To realize this mixed-bit allocation, we \han{use a differentiable bit-width (\ours) that} defines pruning and quantization as a \han{co-optimization} problem, wherein the precision for each SVD component is chosen from a set of integer candidates including zero bits (\textit{e.g.}, within the range [0,5]) \han{in a learnable manner}.
By completely discarding the 0-bit components, we perform rank truncation, which fundamentally acts as parameter pruning. 
Furthermore, our \ours~is inherently free from the strict constraints of maintaining the original structure of the weight matrix.
This is because, unlike unstructured pruning that results in irregular memory layouts~\citep{wen2016learning}, our \ours~naturally preserves hardware-efficient dense computation.
The 0-bit truncation effectively minimizes the inner rank dimension of the decomposed matrices, while maintaining the integrity of dense matrix multiplication.

In addition, to ensure practicality for real-world deployment, our \ours~sorts SVD components into distinct bit-groups. Therefore, we can optimize both memory footprint and computational overhead in mixed-bit-based quantization settings.
This bit-grouping approach enables standardized weight packing and supports weight-quantized general matrix multiplication (GEMM) within each group. 
To validate its practical applicability, we implement a CUDA kernel that performs GEMM operations on our mixed-bit weight matrices, achieving reasonable \han{speedups over} the cuBLAS \han{library}.

We evaluate our \ours~on \han{ten} benchmarks: perplexity in WikiText-2 and C4, and zero-shot accuracy on \han{eight} commonsense reasoning datasets. 
Our experiments cover eight LLMs from the LLaMA-1/2/3 families, spanning parameter scales from 7B to 70B. 
We show that our \ours~achieves \han{favorable} results compared to recent LLM compression approaches, including quantization, structured pruning, SVD, and SVD-based hybrid compression methods under ultra-efficient compression settings.

\section{Method}
\subsection{Motivation: Rethinking Mixed-Bit Allocation}
Before delving into the technical details of the proposed method, we first rethink the current paradigm of mixed-bit allocation in the context of singular value decomposition. 
\han{Precisely}, we seek to answer two fundamental questions: first, an underlying reason for a limited exploration of mixed-bit allocation in prior works; and second, the analytical justification for why our differentiable method is effective.
\paragraph{Why is Mixed-Bit Allocation Underexplored in Existing SVD-based Methods?}
Although SVD-based pruning has shown great promise in LLM compression, mixed-bit allocation has remained largely unexplored in prior works~\citep{yuan2023asvd,wang2024svd}.
The primary bottleneck lies in the conventional reliance on a non-differentiable bit-width approach. 
Prior methods~\citep{huang2024billm,zhao2025ptq1} treat bit-widths as discrete and non-differentiable parameters, which strictly prevents gradient-based optimization.
Therefore, discovering optimal mixed-bit widths presents a significant challenge in terms of time complexity.
Indeed, directly searching for the optimal mixed-bit configuration is intractable, as it requires $O(B^C)$ time for bit candidates $B$ and components $C$.
Since navigating this enormous search space is practically difficult using manual or heuristic approaches, previous SVD-based methods have naturally avoided mixed-bit assignments.
Experimental results supporting our analysis are provided in~\cref{tab:svd_quant}.

\paragraph{Why is Co-optimization of SVD Pruning and Quantization Effective? An Entropy Perspective.}
To overcome the intractable search space, making the bit-widths differentiable proves to be a highly effective solution. 
The core benefit of a differentiable bit-width is that it enables the smooth co-optimization of both weight quantization and rank truncation.
We can analyze the effectiveness of this co-optimization through an entropy perspective.
SVD inherently concentrates the representational energy of a matrix into a small number of dominant singular values, and decomposed matrices naturally produce a low-entropy state with different levels of significance.
Although prior SVD-based methods also utilize this property, they typically perform pruning and quantization sequentially, ultimately applying a \han{uniform-precision} quantization to the surviving components.
This sequential approach is fundamentally incapable of adapting to the various entropy levels within the decomposed matrix. Our differentiable approach directly exploits this entropy variance. By co-optimizing both operations, it adaptively allocates varying mixed bit-widths, assigning zero bits (effective truncation) to less important components and more bits to information-rich ones. This process precisely aligns the quantization resolution with the specific entropy of each of the SVD components. 
Experimental support and entropy-based analyses on this adaptive co-optimization are detailed in \cref{fig:tradeoffcomponent} and~\cref{sec:appendix_singular_value}.

\subsection{Differentiable Bit-Width}
\label{sec:pbl_framework}

\figframework

We introduce a differentiable framework for optimizing bit-widths of SVD components.
Since bit-width parameters are discrete and non-differentiable, directly optimizing them is challenging during backpropagation. 
We enable bit-width updates via \textit{(i) bit-width quantization} and \textit{(ii) block-wise tuning}, \han{allowing gradients to flow to bit-widths (\cref{fig:framework})}.

\textit{(i) Bit-width quantization} serves as a budget-aware algorithm designed to transform floating-point bit-widths into limited integer values.
This aspect is essential because gradients naturally drive all bit-width parameters toward higher precision to reduce the error, which could otherwise easily exceed the given bit budget. To address this, we treat the learnable bit-width parameters as probabilities by normalizing them and allocate the total budget according to these probabilities, ensuring that the overall budget remains constant. 
Specifically, let the decomposed parameters be $\Theta=\text{SVD}(\theta)=\{\mathbf{U}, \Sigma, \mathbf{V}\}$, and denote the $c$-th SVD component as a pair of singular vectors $w_c=\{\mathbf{U}_{:,c}, \mathbf{V}_{:,c}\}$. We then define the continuous bit-width parameters $\beta \in \mathbb{R}^{C}$, where $C$ is the number of SVD components. For a bit-width quantization $\text{BQ}(\cdot)$, we normalize the parameters to [0,1] and allocate a budget $\alpha \in \mathbb{R}$ according to it:
\begin{equation}
    \text{BQ}(\beta_c,\alpha) =  \left\lfloor \alpha \frac{\text{ReLU}(\beta_c)}{\sum_{i=1}^{C} \text{ReLU}(\beta_i)} \right\rfloor,
    \label{eq:bit_quantizer}
\end{equation}
where we use $\text{ReLU}(\cdot)$ to make the bit-widths non-negative and employ the floor operation $\lfloor \cdot \rfloor$ with the straight-through estimator (STE) trick to allow backpropagation in the non-differentiable operation, following the previous work of~\citet{esser2019learned}. However, this floor discretization causes the average of component-level bits to fall short of the budget. 
To strictly satisfy the budget constraint, we distribute the remaining bits to the components with the largest remainders based on Hamilton's method~\citep{balinski1975quota}. A more detailed discussion of this allocation strategy is provided in \cref{sec:appendix_hamilton}. 
Moreover, we employ the learned bit-width quantizer $\mathrm{LBQ}(w_c,\hat{\beta}_c)$ to quantize the SVD component to low precision by using quantized bit-widths $\hat{\beta_c}$. Unlike previous quantizers~\cite{bhalgat2020lsq+}, ours allows the gradients to flow to quantization parameters by employing differentiable operations (see \cref{sec:appendix_lbq} for details).

\textit{(ii) Block-wise tuning} method trains only the target block on a small number of calibration samples (\textit{e.g.}, 4096 samples with sequence length 2048) while freezing the parameters of the other blocks for a fast and memory-efficient learning process. Specifically, we compare the outputs of the quantized parameters, with those of the original ones on the input features $\mathcal{X}$ via the mean squared error ($\mathrm{MSE}$):
\begin{equation}
    L_\text{rec} (\mathcal{X},f,\{\hat{\Theta}^{i},\theta^{i}\}_{i=1}^{M}) = 
    \mathrm{MSE}\!\left(
    f(\mathcal{X};\{\hat{\Theta}^{i}\}_{i=1}^{M}),
    f(\mathcal{X};\{\theta^{i}\}_{i=1}^{M})
    \right).
    \label{eq:recon}
\end{equation}

\algfiggemm

\subsubsection{Bi-level Differentiable Bit-Width}
\label{sec:pbl_bilevel}
We extend our \ours~framework to a bi-level differentiable bit-width scheme with both component-level and matrix-level allocation.
As introduced in the previous section, the component-level bit-widths learn the precision of the SVD components. Similarly, we also optimize the bit budget of each weight matrix for greater flexibility than single-level allocation, while using the same normalization trick to keep the total bit budget constant. Mathematically, we let a bit budget $\alpha^{m} \in \mathbb{R}$ be a learnable parameter for the $m$-th weight matrix within a block, which is initialized as a default matrix-level bit budget $\alpha^{m}_\text{0}$. 
We normalize this parameter to adhere to the total bit budget of the block $T=\sum_{i=1}^{M}\alpha^{i}_\text{0}$.
\begin{equation}
    \mathrm{BA}(\alpha^{m}) = T\frac{\alpha^{m}}{\sum_{i=1}^{M} \alpha^{i}}.
    \label{eq:matrix_level_bit}
\end{equation}
Using~\cref{eq:matrix_level_bit}, our \ours~can smoothly allocate higher bit budgets to a matrix with high-entropy singular values within a total budget (see~\cref{sec:appendix_bilevel} for details).
In implementation details, the matrix-level bit budgets bring a small training time overhead  (\textit{e.g.}, 14 scalar multiplications) and can be saved at inference time. We leave more discussion of bit-width convergence in~\cref{sec:appendix_bit_width}.

\subsubsection{Entropy Minimization Loss}
\label{sec:pbl_entropy}
To foster a truncation-friendly singular-value distribution, we introduce an entropy minimization loss on the normalized singular values. Specifically, we update singular values $\Sigma = (\sigma_1, \dots, \sigma_C)$ via minimizing the entropy loss of the normalized singular-value distribution as
\begin{equation}
L_\text{ent}(\Sigma) =
-\sum_{i=1}^{C} \hat\sigma_i \log(\hat\sigma_i), \;
\text{where} \;\;
\hat{\sigma_{c}} = \frac{\text{ReLU}(\sigma_c)}{\sum_{i=1}^C \text{ReLU}(\sigma_i)}.
\end{equation}
Optimizing this entropy loss encourages a more concentrated singular-value distribution, allowing the matrix to be approximated with fewer SVD components, as discussed in~\cref{sec:appendix_entropy_loss}. Therefore, the final optimization objective is $\arg\min_{\{\Theta^{i},\alpha^i,\beta^i\}_{i=1}^{M}} L_\text{rec} + L_\text{ent}$. Since the learned singular values can be absorbed into the quantization parameters, no additional computation is introduced during inference.  

\tableQuantAcc
\tableQuantPPL
\tableBoth
\figmemory

\subsection{Mixed-precision GEMM}
\label{sec:gemm}

We present a mixed-precision GEMM method to efficiently compute the decomposed matrices compressed into multiple bits. Specifically, our method is twofold: (1) bit-group-wise weight packing and (2) parallel GEMM execution. For (1) bit-group-wise weight packing, since our learning method assigns component-wise mixed precision, we can rearrange the order of SVD components after training to form bit-groups consisting of components with the same bit-widths, as shown in~\cref{fig:gemm}. 
This enables efficient memory packing of weights within the same group, while avoiding irregular layouts.
In the inference phase, packed weights are first loaded from VRAM to cache and then dequantized on the fly, following existing quantization methods~\citep{lin2024awq}. This approach reduces the memory-loading latency incurred during data transfer from VRAM to cache, while the cost of the on-the-fly dequantization amounts to approximately one MAC per element and thus has a negligible impact on the overall computation cost. In terms of (2) parallel GEMM execution, we run multiple small GEMM kernels in parallel, each for a different bit-width group. Each GEMM uses the same input activation and produces a partial output for its assigned bit-group weights, and the final output is obtained by accumulating these partial outputs. 
In practice, launching multiple small GEMM kernels in parallel yields a runtime comparable to that of a single large GEMM kernel. Furthermore, because the proposed method supports 0-bit truncation, it reduces not only memory usage, but also effective FLOPs, achieving speedups in both the memory- and the compute-bound regimes, as shown in~\cref{fig:latency}. We leave the implementation details of the CUDA kernel to~\cref{sec:appendix_bit_packing}.

\section{Experiment}
\subsection{Experimental Setup}

\textbf{Dataset.} Following established evaluation protocols~\citep{shao2023omniquant, zhao2025ptq1}, we assess the performance of the compressed models using two metrics: perplexity and reasoning accuracy. We report perplexity using the WikiText-2~\citep{merity2016pointer} and C4~\citep{raffel2020exploring} datasets. We then evaluate zero-shot accuracy on eight widely used commonsense reasoning benchmarks: PIQA~\citep{bisk2020piqa}, ARC-Easy (ARC-e) and ARC-Challenge (ARC-c)~\citep{clark2018think}, HellaSwag~\citep{zellers2019hellaswag}, WinoGrande~\citep{sakaguchi2020winogrande}, RACE~\citep{lai2017race}, LAMBADA-OpenAI, and LAMBADA-Standard~\citep{paperno2016lambada}. More empirical results of MMLU and GSM8K can be found in~\cref{sec:appendix_zeroshot}.

\noindent\textbf{Calibration.} Our calibration dataset consists of 4096 samples with a sequence length of 2048, which are randomly selected from the RedPajama-1T dataset~\citep{weber2024redpajama}. Following previous practice~\citep{shao2023omniquant,huang2024slim}, our calibration method trains the network in a layer-by-layer manner, which uses the training memory cost similar to the recent post-training quantization (PTQ) baselines. For a fair comparison with the recent extremely low-bit PTQ method~\cite{zhao2025ptq1}, we set the target bit to 1.61 and use a group size of 128. We describe a more detailed hyperparameter setting in~\cref{sec:appendix_experimental_details}.

\noindent\textbf{Baseline methods.} We employ weight-only PTQ methods with a precision of 2-bit or sub-2-bit. For 2-bits, we report the commonly used PTQ baselines such as GPTQ~\citep{frantar2022gptq}, AWQ~\citep{lin2024awq}, QuIP~\citep{chee2023quip}, OmniQuant~\citep{shao2023omniquant}, and SliM-LLM\textsuperscript{+}~\citep{huang2024slim}. Then, we assess the recent sub-2-bit PTQ methods of PB-LLM~\citep{shang2023pb}, BiLLM~\citep{huang2024billm}, and PTQ1.61~\citep{zhao2025ptq1}. Finally, we evaluate the sequential optimization of SVD~\citep{yuan2023asvd,wang2024svd} and quantization methods, and structured pruning~\citep{ma2023llm,chen2024streamlining}.

\subsection{Experimental Results}

\noindent\textbf{Accuracy and perplexity evaluation.} We validate the zero-shot reasoning accuracy of our \ours~against 2-bit and sub-2-bit PTQ baselines in~\cref{tab:zeroshot1} and perplexities on the WikiText-2 and C4 datasets in~\cref{tab:ppl}. We apply our \ours~to LLaMA~\citep{touvron2023llama}, LLaMA-2~\citep{touvron2023llama2}, and LLaMA-3~\citep{dubey2024llama} model families for extensive evaluation. Following the recent practice from~\citep{zhao2025ptq1}, we report the weight bit as well as the mask bit. Since our mixed-bit quantization has nearly zero cost of mask bit (\textit{i.e.}, 4 scalar values per matrix), we denote our mask bit as 0. 
Moreover, because our \ours~is based on the mixed bit-width, we report the bits-per-weight (bpw) to compute the effective bit-width per the original weight. 
\ours~works favorably against the existing 2-bit and sub-2-bit baselines across various model families, indicating that our method preserves the reasoning capability and perplexity of LLMs.
\cref{tab:bitwidth,tab:model} compare PTQ methods across a range of bit-widths and model settings~\citep{team2024gemma,qwen25technicalreport,jiang2023mistral7b}. The proposed method performs consistently well across diverse settings.

\noindent\textbf{Memory-accuracy trade-off evaluation.}
\cref{fig:memory} summarizes the relationship between memory footprint and downstream accuracy for DBW and representative low-bit baselines. While our main study focuses on the LLaMA family, we additionally evaluate Mistral-7B as a non-LLaMA sanity check. DBW remains on a favorable memory-accuracy frontier on all three models, including Mistral-7B, suggesting that the observed advantage is not confined to the LLaMA family. Although this does not constitute an exhaustive cross-architecture study, it provides initial evidence that the proposed co-optimization strategy transfers beyond the specific LLaMA setting.

\figlatency

\noindent\textbf{Latency evaluation.} 
\cref{fig:latency} presents the end-to-end latency comparison of LLaMA-65B using the proposed GEMM against cuBLAS (FP16) and OmniQuant (INT2). For stability, the runtime is the average value of 30 iterations with 10 warm-up repetitions. We detail the setup in \cref{sec:appendix_evaluation_hyperparameters} and discuss it further in \cref{sec:appendix_bit_packing}. For a fair comparison, both OmniQuant and our \ours~are based on the same AWQ~\citep{lin2024awq} implementation of weight-only quantized GEMM. At 64 tokens, \ours~is slightly slower than OmniQuant because the computation per launch is small and the overhead of launching multiple bit-group GEMMs is not fully amortized. As the context length increases, however, the reduced arithmetic and memory movement become the dominant factors, and DBW consistently achieves lower latency than both cuBLAS and OmniQuant.

\noindent\textbf{Hybrid method evaluation.} \cref{tab:svd_quant} assesses the hybrid methods using both the SVD and quantization approaches. For sequential optimization methods, we first perform the pruning with the ASVD or SVD-LLM methods and quantize the remaining weights using the popular 3-bit PTQ methods (\textit{i.e.}, GPTQ, OmniQuant). We set the pruning ratio and bit-width of these methods similarly to \ours, resulting in 1.61 bpw, ensuring an extra recovery step of SVD-LLM following the original work.
Our co-optimization of \ours~performs favorably against the sequential methods under the same bpw.

\tableSVD
\tablepruning

\noindent\textbf{Evaluation of structured pruning method.} 
\cref{tab:pruning} compares our \ours~with two baselines: StreamLine (layer pruning) and LLM-Pruner (channel pruning). For a fair comparison, we set the pruning ratio lower than our \ours~so that the FLOPs of both methods are similar and train the pruned model for recovery, following the original work. Our \ours~works favorably against baselines.

\figtradeoffcomponent

\noindent\textbf{Learned bit-width distribution.} \cref{fig:tradeoffcomponent} shows the component-level bit allocation learned by \ours~on LLaMA-2-7B. For the low-entropy matrix, where singular values are concentrated in a few leading components, \ours~keeps only these dominant components and prunes most tail components. For the high-entropy matrix, where singular values decay more gradually, \ours~retains positive bits over a broader component range. This shows that \ours~adapts the bit allocation to each matrix's singular-value spectrum instead of using a fixed truncation threshold.

\tableFactorAnalysis

\noindent\textbf{Controlled factor analysis.} 
Table~\ref{tab:factor} separates the effects of the SVD-space bit-allocation rule and the learnable calibration targets. In Table~\ref{tab:factor1}, Uniform is used mainly as a non-adaptive baseline under the 1.61-bit budget, while Heuristic provides a stronger non-differentiable, SVD-aware reference based on singular-value importance. The learnable variant further improves over this discrete allocation, suggesting that differentiable bit learning brings additional benefits beyond static SVD-aware assignment. Table~\ref{tab:factor2} shows that component-level bit learning is already effective, and that adding matrix-level budget learning and singular-value adaptation further refines the truncation-quantization trade-off across heterogeneous projection matrices.

\vspace{-0.4em}
\section{Related Work}
\vspace{-0.4em}

\noindent\textbf{Post-training Quantization.} 
Recent studies~\citep{frantar2022gptq,xiao2023smoothquant,shao2023omniquant,lin2024awq} propose post-training methods for quantizing LLMs to low bit-widths with modest computational overhead.
~\citet{chen2025efficientqat} further introduce a block-wise QAT approach, in which all other layers are fixed and each block is optimized sequentially, reducing memory cost while requiring only a small calibration set.

\vspace{-0.2em}

\noindent\textbf{Mixed-bit Quantization.}
The unstructured mixed-bit quantization method~\citep{shang2023pb} assigns a high bit for the salient weights while setting a low bit for the remaining weights for ultra-efficiency. They recover accuracy by using an efficient post-training process (\textit{e.g.}, GPTQ or QAT). Similarly, \citet{huang2024billm} define Hessian-based importance to select a prominent weight, which works favorably compared to the existing criterion. \citet{zhao2025ptq1} propose a column-wise mixed-bit quantization method to lower the storage requirements of bit masks in unstructured approaches.

\vspace{-0.2em}

\noindent\textbf{Singular Value Decomposition.}
The main advantage of SVD is that it reduces both memory usage and computational cost simultaneously, which is particularly beneficial for LLM acceleration. \citep{yuan2023asvd} incorporate activation statistics into the decomposition process to preserve the most prominent channels, while \citep{wang2024svd} use a whitening matrix to better align the truncation loss. Several studies~\citep{yuan2023asvd,wang2024svd} have explored integrating pruning and quantization through SVD and reported promising results. However, they treats pruning and quantization separately, as discussed further in Appendix of~\cref{sec:appendix_related_work}.

\vspace{-0.4em}
\section{Conclusion}
\vspace{-0.4em}
We study a hybrid compression approach that combines pruning and quantization to achieve an effective trade-off in singular value decomposition. We propose an integrated framework that unifies the two techniques as a single bit-allocation problem, where each SVD component is assigned a precision selected from zero-inclusive integer bit candidates. To make the framework practical, we further develop a differentiable bit-width method that learns this trade-off in a gradient-based manner and generalizes well across diverse singular-value distributions. Experimental results show that our \ours~framework performs favorably against existing compression methods, particularly in the ultra-efficient regime. We hope that this work provides a foundation for subsequent research devoted to compressing large-scale models to an ultra-small scale. 
\textbf{Limitation.} 
Our evaluation is currently limited to language models. Although we focus on pre-trained LLMs under constrained computational resources to support edge deployment, future work should further validate the generality of our framework across vision, speech, and multimodal architectures.

\section*{Acknowledgments}
This research was supported by the National Research Foundation of Korea (NRF), Electronics and Telecommunications Research Institute (ETRI), and Institute of Information \& Communications Technology Planning \& Evaluation (IITP), funded by the Korean government [26CS1100, Development of Proprietary Physical AI-based Small-scale Computers and Integrated Soft Suits], the Korean government (MSIT), and the Ministry of Education (RS-2024-00356486, RS-2026-25617480, IITP-2026-RS-2023-00255968, and RS-2025-25424589).

\medskip
{
\small
\bibliographystyle{ieeenat_fullname}
\bibliography{egbib}
}

\clearpage
\appendix

\section{Experimental Details}
\label{sec:appendix_experimental_details}
\subsection{Calibration Hyperparameters}
\label{sec:appendix_hyperparameters}
\cref{tab:hyperparameter} provides a detailed hyperparameter specification used in our experiments. Our calibration set comprises 4096 samples with a sequence length of 2048, randomly drawn from RedPajama-1T~\citep{weber2024redpajama}, which uses a comparable number of samples during training against the recent PTQ baselines as shown in~\cref{tab:calibration}. The calibration samples are tokenized using the default HuggingFace\footnote{\texttt{transformers==4.40.1}} tokenizer of each model. Following prior practice~\citep{shao2023omniquant,huang2024slim}, we perform block-wise calibration, training the network one block at a time to keep the memory footprint comparable to recent PTQ baselines. Due to the efficient memory usage of the proposed method, we can calibrate the LLaMA-1, 2, and 3 model families using the RTX 3090 GPUs with 24GB of VRAM and the Xeon Gold CPU with 256GB of RAM. Each block is trained in a batch size of 2 for two epochs using the Adam optimizer. We choose the learning rate of 2e-5 for the weight parameter, including the projection and normalization layer. After this calibration, we perform a quick quantization hyperparameter calibration to reduce the distribution shift among different blocks, following the recent works~\citep{chen2025efficientqat}. For a fair comparison with an extremely low-bit PTQ method~\citep{zhao2025ptq1}, we set the target bit-width to 1.61 and use a group size of 128. 

\tableHyperparameter
\tablecalibration

\subsection{Evaluation Hyperparameters}
\label{sec:appendix_evaluation_hyperparameters}
This subsection provides hyperparameters used to evaluate our \ours~by the perplexity, accuracy, bits-per-weight, weight memory footprint, and runtime. For the perplexity, we use the validation samples of WikiText-2 and C4 with the sequence length of 2048, following the prior work of \citet{frantar2022gptq}. For accuracy, we use the widely used \texttt{lm\_eval}\footnote{\texttt{lm\_eval==0.4.2}} library to assess the commonsense reasoning accuracy of our \ours~and adopt the \texttt{acc} value as an evaluation metric following the previous works of \citet{zhao2025ptq1}. Additionally, we report the bits-per-weight (bpw) to compute the effective bit-width of original weights, which is a standard metric to compute the mixed bit-widths in the existing works. 

Similarly, the weight memory footprint is computed by summing the memory required for the embedding and non-embedding layers. Since the compression method is applied only to the non-embedding layers, the overall weight memory footprint achieves a lower compression ratio than the non-embedding layers. Furthermore, we include the quantization hyperparameters when computing the memory footprint of the non-embedding layers to show the practical memory usage when the model is loaded onto GPUs. These quantization hyperparameters make the quantization compression ratio lower than that of truncation. Consequently, due to the previously mentioned two reasons, the model-level compression ratio is smaller than the theoretical ratio implied by the bits-per-weight of $16 / \text{e-bit}$ (\textit{e.g.}, the bits-per-weight of 1.61 would suggest $10\times$ reduction as $16 / 1.61 \sim 10$). In contrast, the gap between theoretical and practical weight memory savings is evident not only in our \ours~but also in the baseline. Therefore, the difference between the theoretical and empirical compression ratios reflect inherent system-level overheads rather than an algorithmic constraint of our approach.

To evaluate the runtime of the CUDA GEMM kernel, we use the NVIDIA RTX 3090 GPU with 24GB of VRAM. We use the \texttt{torch} library to call the official cuBLAS utilizing the Tensor Core and employ AWQ implementation~\citep {lin2024awq} for OmniQuant and our \ours~for a weight-only quantized GEMM. For stable results, we report the averaged runtime (ms) for 30 iterations with 10 warm-up repetitions.

\subsection{Our \ours~Hyperparameters}
Since our \ours~introduces two types of learnable parameters (\textit{i.e.}, matrix-level and component-level bit-width) to co-optimize pruning and quantization in a single step, we must specify their learning rates. For simplicity, we use the same learning-rate setting across all target LLMs. Specifically, we set the component-level learning rate to 2e-5, following the default learning rate used for weight optimization, and the matrix-level learning rate to 1e-4 to encourage faster convergence compared to other learnable parameters. As our primary goal is to demonstrate the effectiveness of the differentiable bit-width learning framework, other than the effect of optimal hyperparameter choices, we adopt a single learning-rate configuration that generalizes across a wide range of model sizes, thereby minimizing the influence of model-specific hyperparameter tuning. However, in practical applications, more carefully tuned learning rates for each model could further improve performance.

\subsection{Statistical Stability}
\label{sec:appendix_statistic}

\tablestat

The main tables of manuscript report a single run result with a random seed of 46.
To assess the robustness of our results to calibration randomness, we repeat the LLaMA-2-7B experiment at 1.61 bits using three different calibration seeds. 
\cref{tab:statistical_variation} demonstrates the mean, standard deviation, and 95\% confidence interval for both perplexity and accuracy. 
The observed variation is small for both metrics and remains substantially smaller than the performance gaps between methods reported in the main results. 
This indicates that the reported performance is stable across calibration seeds and is not driven by a particular random seed.

\section{Extended Related Work}
\label{sec:appendix_related_work}
\subsection{Hybrid Compression}
Previously, a few works studied the hybrid methods that combine two different compression approaches. As discussed in the manuscript, a mix of truncation and quantization performs well in the ultra-efficient compression ratio. In addition to the already introduced approach, there are two extra branches of hybrid compression methods: (1) unstructured pruning with quantization, and (2) structured pruning with quantization. The representative example of (1) unstructured pruning with quantization~\citep{mishra2021accelerating} is to apply quantization on the already pruned weight, which, however, includes a complicated design due to the unstructured pruning to accelerate the compressed model on the GPUs. Similarly, (2) structured pruning with quantization~\citep{zhou2025qpruner} also applies the structured pruning and then performs a quantization on it. This approach can have a practical speed gain from the modern GPU; however, it loses substantial information in the first pruning step when the pruning ratio is higher than 20\%. 

\subsection{Unified Mixed-bit Quantization}
\label{sec:appendix_rel_unified_mixed_bit}
Prior work on unified mixed-bit quantization that includes 0-bit as a candidate bit-width is limited. \citet{van2020bayesian} propose a bit-width learning approach that includes 0-bit as a candidate. In theory, this method can learn the optimal trade-off between quantization and pruning. However, there are four limitations that prevent this method from being directly applied to ultra-efficient LLM compression. First, this method brings a considerable training overhead according to the length of bit candidates, which is moderate in a lightweight vision CNN; however, it becomes prohibitive in an LLM with more than $100\times$ as many parameters as the CNN. Therefore, the existing method is difficult to use to extremely compress current LLMs that use a billion-scale transformer-based architecture. Furthermore, it needs to repeatedly retrain the model to find an exact bits-per-weight. Since there is no regularization of bit budget that keeps the average learned bit under the budget, we need a full hyperparameter search to achieve the specific target bit-width, which can lead to our training cost being intractable for compressing LLMs. Moreover, since this method selects bit-widths from the limited candidates such as 0, 2, 4, and 8, it is difficult to find an optimal trade-off under the ultra-efficient compression ratio. Finally, they perform channel-pruning at the original weight matrix and remove whole channels from the forward pass if the learned bit is selected as a 0-bit, which could disrupt the whole computation flow and yield unexpected results in practical transformer architectures that LLMs are usually built on. 

\section{Extended Discussion}

\subsection{Optimal Trade-off: SVD Entropy View}
\label{sec:appendix_singular_value}
\figoptimaltradeoff
Finding an optimal trade-off between truncation and quantization is important on the decomposed matrices due to the high variability of SVD entropy (\textit{i.e.}, entropy of normalized singular values)~\citep{roy2007effective}. The SVD entropy represents whether information of the matrix is concentrated in a few dominant singular values or more evenly distributed, which determines which method is more appropriate for the target matrix between truncation and quantization. In a low SVD entropy of \cref{fig:optimaltradeoff_2}, quantization results in a large reconstruction error since it applies a uniform precision across components, which allocates insufficient precision to the dominant ones. However, in a high SVD entropy of \cref{fig:optimaltradeoff_3}, truncation inevitably removes notable information from the uniform distribution, yielding suboptimal results. Our goal is to adaptively allocate the trade-off ratio to minimize the error across a wide range of SVD entropy.

As shown in~\cref{fig:singularvalue}, this entropy-dependent trade-off can be interpreted as a bit-allocation problem over SVD components. A static allocation rule relies on fixed, non-differentiable thresholds, and therefore applies the same precision boundary regardless of how the singular values are distributed. In contrast, our \ours~learns component-wise bit-widths through the reconstruction objective, allowing the effective thresholds to adapt to each matrix. This adaptive allocation assigns zero bits to components that can be safely removed and positive bits to components that should be retained, thereby realizing pruning and quantization within a single learnable rule. We provide empirical analyses of the learned SVD-entropy-dependent allocation in~\cref{sec:appendix_optimal_tradeoff}.
\figsingularvalue

\subsection{Hamilton Method}
\label{sec:appendix_hamilton}
\tableHamilton
Our component-level bit quantization allocates the matrix bit budget to components according to the component-level bit-width parameter distribution. For this, we use the two-round allocation method. In the first round, we assign the matrix bit budget to components according to the component-level bit-width parameter distribution and apply the floor operation to make them integer bit-width. This floor operation results in the leftover bit $R$ since it truncates the remainder $r_i$. 
\begin{equation}
    R_c = \sum_{i=1}^{C} r_{i} = \sum_{i=1}^{C} \beta_{i} - \hat{\beta}_{i},
\end{equation}
where $\hat{\beta}_{i}$ means the quantized component-level bit-width, introduced in the manuscript of~\cref{eq:bit_quantizer}.
In the second round, we use Hamilton's method~\citep{balinski1975quota} to allocate the leftover bit budget to components. This method uses the largest remainder approach that select the top-$R$ components with the largest remainder as $\tau := \mathrm{top}\left(\{r_{i}\}_{i=1}^{C}, R\right)$ and assign the leftover bit to them, making the bit-width results to resemble the original component-level distribution.
\begin{equation}
    r_{i} \leftarrow r_{i} + 1 \;\; \forall i \in \tau.
\end{equation}
By using this two-round allocation with Hamilton's method, we can fully allocate the matrix budget to components, following the component-level bit-width as closely as possible. 
In sort, the first round allows the gradient decides which components matter, and the second round only projects that to the integer budget.

Furthermore, we want to discuss an important topic regarding the gradient mismatch introduced by Hamilton’s method of our bit-width quantizer. To clarify, let $\beta_c$ be the continuous bit-width and $\hat{\beta_c} = \text{BQ}(\beta_c,\alpha) + h$ be the final discrete bit-width, where $h \in \{0, 1\}$ represents the bit adjustment assigned by Hamilton’s method. During the backward pass, we employ the straight-through estimator by setting the gradient $\frac{\partial L}{\partial \beta_c} \approx \frac{\partial L}{\partial \hat{\beta_c}}$. Specifically, the sorting and top-k allocation step is treated as a non-differentiable constant offset, meaning we assume $\frac{\partial h}{\partial \beta_c} = 0$. While this inevitably introduces a slight gradient mismatch, the magnitude of this error is strictly bounded to at most $1$ bit per affected component. In the context of stochastic gradient descent, this bounded mismatch effectively acts as localized quantization noise. Since the primary continuous parameter $\beta_c$ still receives the correct global directional gradient from the loss landscape, the overall optimization trajectory remains stable and converges to a robust solution, as demonstrated by our empirical results in the manuscript. 

We further examine whether the non-differentiable Hamilton correction introduces a measurable optimization mismatch. After the continuous bit allocation is floored, the remaining bit budget must be assigned to a subset of components through the binary indicator $h_i$. \cref{tab:hamilton} compares three forward allocation rules for this indicator while keeping the same continuous bit-width parameters and total matrix-level budget. The Floor-only variant sets $h_i=0$ for all components and therefore removes the remainder-assignment step. This variant no longer preserves the target bit budget after flooring, which makes it an unsuitable projection for strict-budget calibration. The Random variant preserves the number of assigned leftover bits, but chooses the components randomly, introducing stochastic discontinuities that are unrelated to the continuous bit-width distribution.
In contrast, Hamilton's method assigns the leftover bits to the components with the largest remainders:
\[
h_i =
\begin{cases}
1, & i \in \gamma, \\
0, & i \notin \gamma,
\end{cases}
\]
where $\gamma$ denotes the set of components selected by the largest-remainder rule.
This projection exactly preserves the target budget while minimizing the rounding discrepancy from the continuous allocation.
The stable performance of the Hamilton variant indicates that the discrete correction does not introduce a harmful optimization mismatch in practice.

\subsection{Learned Bit-Width Quantization}
\label{sec:appendix_lbq}

\figlbq

The proposed learned bit-width quantization explicitly uses the integer bit-width parameter in quantizing the weight as shown in~\cref{fig:lbq}. The existing quantizer (\textit{i.e.}, learned step size quantizer (LSQ)) employs a bit-width parameter as an argument to the clipping operation that sets the maximum values of weights. However, the gradient of the clipping operation is somewhat limited to learn the optimal bit-width. To solve this problem, we use the bit-width parameter as a scaling factor, which provides a more dynamic learning signal to them. Precisely, the baseline LSQ transform a pair of singular vectors $w_c \in \{U_{:,c}, V_{:,c}\}$ into low-bit $\hat{\beta_c}$ with learnable step size $s$ and zero point $z$:
\begin{equation}
\mathrm{LSQ}(w_c,s,z,\hat{\beta}_c) = \left\lfloor\text{clip}(w_c/s-z, 0, 2^{\hat{\beta_c}}-1)\right\rceil s + z.
\label{eq:lsq}
\end{equation}
Through~\cref{eq:lsq}, LSQ aims to learn the optimal step size and zero point to minimize the quantization error. Unfortunately, since the clipping operation yields a limited gradient to the bit parameter, this method does not provide a meaningful learning signal for the bit parameters. To address this issue, we introduce a learned bit-width quantizer $\mathrm{LBQ}(\cdot)$ that converts a pair of singular vectors into low precision. Specifically, we employ min--max normalization $N(\cdot)$ to transform them into the range $[0,1]$ and scale the normalized values according to the unsigned integer range $[0, 2^{\hat{\beta}_c}-1]$ as
\begin{equation}
\mathrm{LBQ}(w_c,\hat{\beta}_c)
=
\begin{cases}
\mathbf{0}, & \hat{\beta}_c = 0, \\[4pt]
N_{\mathrm{mm}}^{-1}\!\left(
\displaystyle
\frac{
\left\lfloor N(w_c)\,(2^{\hat{\beta}_c}-1)\right\rceil
}{
2^{\hat{\beta}_c}-1
}
\right),
& \hat{\beta}_c > 0,
\end{cases}
\label{eq:q}
\end{equation}
where $\left\lfloor\cdot\right\rceil$ denotes the round-to-nearest operation and $N_{\mathrm{mm}}^{-1}(\cdot)$ denotes the inverse of min--max normalization. When $\hat{\beta}_c=0$, the factor $2^{\hat{\beta}_c}-1$ vanishes in both the numerator and denominator, making the quantization expression undefined. We therefore define the output as zero for this case, since a zero-bit component is removed rather than quantized.

\figlbqgradient

As shown in~\cref{fig:lbq_gradient}, \cref{eq:q} provides a learning signal to a bit parameter to choose adequate values from the backpropagation, unlike the baseline LSQ that flows the gradient to bit-width if and only if the weights are out of the maximum range.
Practically, in inference phase, the dequantization process of~\cref{eq:q} is fast since it can be reduced into a simple multiply-and-add operation $sw_\text{int} + z$ where $s=\frac{w_\text{max} - w_\text{min}}{2^{\hat{\beta}}-1}$, and $z=w_\text{min}$.

\subsection{Matrix-level 0-bit Collapse}
\label{sec:appendix_bit_width}
\figbittrajectory
We propose a differentiable bit-width learning method that jointly optimizes matrix-level and component-level precision for the decomposed matrices.
From a convergence perspective, the main concern when training the bit-width parameters is preventing the matrix-level bit-width from collapsing to zero, which can lead to unexpected behavior in current transformer architectures that have complex feature connections in the attention and MLP block. In theory, such a collapse could occur because the matrix-level bit-width in~\cref{eq:matrix_level_bit} does not include an explicit regularizer that prevents the bit-width from becoming zero. However, in practice, as shown in~\cref{fig:trajectory}, we observe that the matrix-level bit budgets never collapse to zero under our current formulation. This is because the reconstruction loss increases drastically when a specific matrix's bit budget approaches zero. Consequently, the loss inherently acts as an implicit lower-bound regularizer, effectively preventing the bit budget parameters from becoming zero. Therefore, our \ours~chooses not to use an additional regularization for this problem and converges favorably to an optimal trade-off without suffering from matrix-level 0-bit collapse. 
This behavior is an empirical observation of the loss landscape and does not constitute a theoretical guarantee.

\subsection{CUDA Kernel Optimization}
\label{sec:appendix_bit_packing}
\figdequant
\noindent\textbf{Bit packing.} Since our \ours~employs bit-width from 0 to 5, we use two different bit-packing schemes depending on whether the bit-width can evenly pack into a 32-bit word: (1) evenly packable and (2) non-evenly packable bit-widths. For (1) evenly packable bit-widths, we can easily use the widely used packing approach~\cite{lin2024awq} to pack weight with the power of 2 bits since the commonly used 32-bit word is divisible by 1, 2, and 4 bits. However, for (2) non-evenly packable bit-widths, directly saving them results in extra memory overhead because they lead to leftover bits in a 32-bit word. Therefore, we need to design a packing approach that can save the weight into non-divisible bit-widths without extra memory overhead. For this, we employ a divide-and-conquer strategy that saves a weight in separate bit-widths that can evenly pack into a 32-bit word (\textit{e.g.}, 3-bit into 2 and 1-bit) and concatenate them via our on-the-fly dequantization kernel where we can utilize a fast shift and $\&$ operator as shown in~\cref{fig:dequant}, which introduce a small overhead from the baseline dequantization computation cost of $\mathcal{O}(n)$ to $\mathcal{O}(1.5n)$ regarding to the matrix size $n$ that is still negligible compared to ones of GEMM operation $\mathcal{O}(nT)$ when the token size $T$ is large enough. 
Consequently, this divide-and-conquer method achieves the theoretical memory compression ratio for non-dividable bit-width as $16 / \text{bit}$ while being fast to load weight into memory since the adding one more item of extra weight into a source memory address list that has more than three (e.g., weight, scale, and zero-point) does not introduce the memory access bottleneck and the concatenation overhead is negligible in the efficient CUDA kernel (see~\cref{tab:tablebitpack}). 

\tablebitpack

\noindent\textbf{Parallel GEMM execution.} We decompose a large GEMM kernel into multiple smaller GEMMs with different bit-widths. To execute these kernels in parallel, we use CUDA streams, which allow multiple kernels to run concurrently through the official CUDA API. We adopt this straightforward implementation to focus on validating the effectiveness of the proposed mixed-precision GEMM, rather than optimizing kernel-level performance. However, developing a fully fused CUDA kernel could further reduce kernel launch overhead and improve runtime efficiency. Therefore, in future research, we believe that researchers and practitioners interested in production-level deployment can achieve additional speed improvements by implementing such fused kernels.

\subsection{Hyperparameter Sensitivity}

\tablegroupsize

We use a group size of 128 to match the sub-2-bit baselines (OmniQuant, SliM-LLM\textsuperscript{+}, and BiLLM) used in our comparison. To examine the sensitivity to this choice, we additionally evaluate group sizes of 128 and 256 on LLaMA-2-7B. As shown in Table~\ref{tab:group_size}, our method remains stable: the perplexity changes only from 10.54 to 11.02, while the accuracy changes from 49.06 to 47.72. In comparison, BiLLM's perplexity increases from 35.60 to 39.55 over the same group sizes.

\section{Extended Experimental Results}
\subsection{Instruction-tuned Model Evaluation}

\tableinstruct

To evaluate whether our method generalizes beyond base pretrained models, we additionally test it on instruction-tuned, distilled, and chat variants of LLaMA. 
These models have different training objectives and response behaviors from the base models used in our main experiments, making them a useful test of whether the quantization method remains reliable after post-training. 
We apply the same quantization setting across all models and compare against existing low-bit baselines under the same group size. 
Overall, our method favorably preserves both perplexity and downstream accuracy against the competing methods, as shown in~\cref{tab:instruct_models}. 
This suggests that our approach is not limited to a particular model variant and remains effective across different forms of post-trained LLMs.

\subsection{Zero-shot Reasoning Benchmarks}
\label{sec:appendix_zeroshot}

\tableQuantAccTwo

\cref{tab:zeroshot2} evaluates the proposed method against an ultra-efficient quantization baseline in terms of zero-shot reasoning accuracy on the MMLU and GSM8K benchmarks. These benchmarks are widely used to assess complex reasoning and multi-step problem solving, making them particularly sensitive to aggressive compression. As discussed in prior work~\citep{xiao2025task}, LLMs operating in ultra-efficient quantization regimes often suffer substantial accuracy degradation on MMLU and GSM8K, where preserving subtle numerical and logical cues is crucial. In this setting, our \ours~consistently performs well against the baseline, indicating that the proposed trade-off between truncation and quantization better preserves reasoning capability under tight memory constraints. Nevertheless, the absolute performance gap from the full-precision model remains non-negligible, suggesting that further improvements are still needed to fully close the accuracy drop in ultra-low-bit regimes.

\subsection{Sequential vs. Co-optimization}
\label{sec:appendix_sequential_hyperparameter_search}
\tableSequentialSearch
\cref{tab:svd_quant} of the manuscript compares the sequential optimization against our co-optimization method on three different LLaMA models. Going one step further, \cref{tab:tableSequentialSearch} extends the evaluation target to different bit-widths. This evaluation performs a considerable hyperparameter search (\textit{e.g.}, pruning ratio, quantization bit-width for remaining components, and quantization method) to show optimal results of sequential optimization using SVD-based pruning and quantization, while fixing SVD-LLM as the pruning method since we observe a good performance gain of SVD-LLM compared to the ASVD approach. Our \ours~performs well against the sequential optimization method equipped with a moderate hyperparameter search, except in the case of 0.5 bit-width, where sequential optimization performs moderately against ours by using an extra amount of memory. This confirms that our initial argument for co-optimization is supported by a fair amount of empirical evidence.

\section{Extended Analysis}
\label{sec:appendix_analysis}

\tabletradeoff
\figtradeoffblock
\tabletradeofftwo

\subsection{Optimal Trade-off}
\label{sec:appendix_optimal_tradeoff}
We analyze the truncation-quantization trade-off learned by \ours~at three different granularity: model-level, block-level, and matrix-level. This multi-level analysis allows us to examine how \ours~distributes the compression budget globally across the model, locally across transformer blocks, and specifically across individual projection matrices.

\noindent\textbf{Model-level} \cref{tab:tradeoff} summarizes the model-level truncation-quantization trade-off learned by \ours. Across the LLaMA family, \ours~prunes $42.91\%-48.90\%$ of SVD components and assigns an average precision of $2.82$-$3.15$ bits to the remaining components. This shows that \ours~does not rely solely on either aggressive truncation or extremely low-bit quantization; instead, it combines a moderate pruning ratio with a reasonably higher precision for the retained components. As a result, \ours~achieves up to $9\times$ weight-memory compression while maintaining the compressed model in a practical memory range. Specifically, the effective model-level compression ratio becomes larger for larger models. This is because DBW is applied to the non-embedding projection matrices, whereas embedding and normalization layers are kept in higher precision. In smaller models, these uncompressed components occupy a relatively larger portion of the total model memory, reducing the observed model-level compression ratio. In larger models, most parameters are concentrated in the transformer projection matrices, so the benefit of compressing non-embedding layers is reflected more directly in the total memory footprint.

\tableBiLevelAnalysis

\noindent\textbf{Block-level} \cref{fig:tradeoffblock} provides a block-level view of the truncation-quantization trade-off learned by \ours. As shown in \cref{fig:tradeoffblock_1}, MLP blocks generally have higher SVD entropy than attention blocks, indicating that their singular-value energy is more evenly distributed across components. Because truncating high-entropy matrices can remove a larger amount of distributed information, \ours~assigns a lower pruning ratio to MLP blocks and a higher pruning ratio to attention blocks, as shown in \cref{fig:tradeoffblock_2}. Instead of relying only on truncation, \ours~allocates more quantization capacity to the MLP blocks: the remaining components in MLP receive slightly higher bit-widths in \cref{fig:tradeoffblock_3}, and the resulting bits-per-weight are also higher than those of attention in \cref{fig:tradeoffblock_4}. These trends show that \ours~does not apply a uniform compression rule across blocks, but adaptively balances rank truncation and quantization according to the SVD characteristics of each module.

\noindent\textbf{Matrix-level} In addition to block-level analysis, \cref{tab:tradeofftwo} observes a similar trend in the matrix-level analysis, which indicates that attention projection matrices generally have lower SVD entropy than MLP matrices, and thus receive higher truncation ratios. Also, within the same micro-block (attention, MLP), the matrix-level bit budgets (\textit{Bit}) have a moderate variation according to their SVD entropy. This empirical evidence supports our hypothesis that our \ours~can find an appropriate trade-off even when SVD entropy varies substantially across weight matrices and distribute block bit budget to matrices according to their SVD entropy. 

\subsection{Bi-level Differentiable Bit-Width}
\label{sec:appendix_bilevel}
\cref{tab:bilevel} shows that our algorithm yields more flexible bit-width by allowing moderate differences in the bits-per-weight and memory compression ratios within a block. Instead of imposing a strictly uniform quantization pattern, this design provides additional freedom to adapt the allocation to the local characteristics of the parameters in each block. Such flexibility leads to more efficient use of the available bit budget while better preserving the information required for model performance.

\subsection{Singular Value Entropy Minimization}
\label{sec:appendix_entropy_loss}
\figsvdentropy
We propose a minimization loss designed to reduce the entropy of singular values during training for LLM compression. As shown in \cref{fig:svdentropy}, this loss effectively encourages the singular-value distribution to become more concentrated, allowing the model information to be represented by a smaller number of dominant SVD components compared with the baseline SVD approach. In other words, our method promotes a more compact representation by increasing the relative importance of a few major components while suppressing less important ones. In addition, our method drives SVD components associated with small singular values closer to zero. This behavior is especially beneficial during inference, because it enables more aggressive truncation of low-rank components than the original SVD without severely affecting the representation quality. As a result, the proposed method provides a more compression-friendly singular-value structure, which can improve the efficiency of LLM deployment.

\section{Extended Limitation and Future Direction} 

We discuss the limitations of this work and the promising directions for the future research.

\begin{itemize}
    \item Our training procedure is performed block-wise for efficiency, so the optimization does not explicitly capture global interactions across all layers. Extending the method to end-to-end global training is a natural direction for future work.
    \item Our study targets the sub-2-bit regime, where preserving model quality under extreme compression is particularly challenging.
    \item Our study focuses on weight quantization. We do not consider activation quantization or KV-cache quantization, both of which can further affect memory usage and inference efficiency in practical deployment. Extending the framework to jointly account for these components is an important direction for future work.
\end{itemize}

\newpage
\section*{NeurIPS Paper Checklist}

\begin{enumerate}

\item {\bf Claims}
    \item[] Question: Do the main claims made in the abstract and introduction accurately reflect the paper's contributions and scope?
    \item[] Answer: \answerYes{} 
    \item[] Justification: The abstract and introduction clearly state our main claims that the proposed differentiable bit-width facilitates co-optimizing pruning with quantization. 

\item {\bf Limitations}
    \item[] Question: Does the paper discuss the limitations of the work performed by the authors?
    \item[] Answer: \answerYes{} 
    \item[] Justification: We create a separate ``Limitations'' section in our paper pointing out that our assumptions should be validated in other domains.

\item {\bf Theory assumptions and proofs}
    \item[] Question: For each theoretical result, does the paper provide the full set of assumptions and a complete (and correct) proof?
    \item[] Answer: \answerNA{} 
    \item[] Justification: Our paper focuses on empirical results.

    \item {\bf Experimental result reproducibility}
    \item[] Question: Does the paper fully disclose all the information needed to reproduce the main experimental results of the paper to the extent that it affects the main claims and/or conclusions of the paper (regardless of whether the code and data are provided or not)?
    \item[] Answer: \answerYes{} 
    \item[] Justification: In experimental setting section of manuscript and experimental detail section of appendix, we provide all the information required to reproduce our experiments in details.

\item {\bf Open access to data and code}
    \item[] Question: Does the paper provide open access to the data and code, with sufficient instructions to faithfully reproduce the main experimental results, as described in supplemental material?
    \item[] Answer: \answerYes{}
    \item[] Justification: We publicly release our code. The link is provided in the abstract.

\item {\bf Experimental setting/details}
    \item[] Question: Does the paper specify all the training and test details (e.g., data splits, hyperparameters, how they were chosen, type of optimizer) necessary to understand the results?
    \item[] Answer: \answerYes{} 
    \item[] Justification: In experimental setting section of manuscript and experimental detail section of appendix, we provide all the training and test details.

\item {\bf Experiment statistical significance}
    \item[] Question: Does the paper report error bars suitably and correctly defined or other appropriate information about the statistical significance of the experiments?
    \item[] Answer: \answerYes{} 
    \item[] Justification: We provide statistical significance in~\cref{sec:appendix_statistic}.

\item {\bf Experiments compute resources}
    \item[] Question: For each experiment, does the paper provide sufficient information on the computer resources (type of compute workers, memory, time of execution) needed to reproduce the experiments?
    \item[] Answer: \answerYes{} 
    \item[] Justification: In experimental detail section of appendix, we provide all the information on the computer resources used to run our experiments in details.
    
\item {\bf Code of ethics}
    \item[] Question: Does the research conducted in the paper conform, in every respect, with the NeurIPS Code of Ethics \url{https://neurips.cc/public/EthicsGuidelines}?
    \item[] Answer: \answerYes{} 
    \item[] Justification: We review the NeurIPS Code of Ethics thoroughly and try to keep these rules.

\item {\bf Broader impacts}
    \item[] Question: Does the paper discuss both potential positive societal impacts and negative societal impacts of the work performed?
    \item[] Answer: \answerNA{} 
    \item[] Justification: Our work is to compress the LLMs into ultra-efficient format, which help to reduce their computational resources. Since this work is foundational research and not tied to particular application, we believe that this work has no societal impact.
    
\item {\bf Safeguards}
    \item[] Question: Does the paper describe safeguards that have been put in place for responsible release of data or models that have a high risk for misuse (e.g., pre-trained language models, image generators, or scraped datasets)?
    \item[] Answer: \answerNA{} 
    \item[] Justification: Our main goal is to reduce the computational cost of LLMs. Therefore, our project does not release a new model or dataset.

\item {\bf Licenses for existing assets}
    \item[] Question: Are the creators or original owners of assets (e.g., code, data, models), used in the paper, properly credited and are the license and terms of use explicitly mentioned and properly respected?
    \item[] Answer: \answerYes{} 
    \item[] Justification: We cite the original paper that produced the code package or dataset.

\item {\bf New assets}
    \item[] Question: Are new assets introduced in the paper well documented and is the documentation provided alongside the assets?
    \item[] Answer: \answerNA{} 
    \item[] Justification: Our paper does not release new assets.

\item {\bf Crowdsourcing and research with human subjects}
    \item[] Question: For crowdsourcing experiments and research with human subjects, does the paper include the full text of instructions given to participants and screenshots, if applicable, as well as details about compensation (if any)? 
    \item[] Answer: \answerNA{} 
    \item[] Justification: Our paper does not involve crowdsourcing nor research with human subjects

\item {\bf Institutional review board (IRB) approvals or equivalent for research with human subjects}
    \item[] Question: Does the paper describe potential risks incurred by study participants, whether such risks were disclosed to the subjects, and whether Institutional Review Board (IRB) approvals (or an equivalent approval/review based on the requirements of your country or institution) were obtained?
    \item[] Answer: \answerNA{} 
    \item[] Justification: Our paper does not involve crowdsourcing nor research with human subjects.

\item {\bf Declaration of LLM usage}
    \item[] Question: Does the paper describe the usage of LLMs if it is an important, original, or non-standard component of the core methods in this research? Note that if the LLM is used only for writing, editing, or formatting purposes and does \emph{not} impact the core methodology, scientific rigor, or originality of the research, declaration is not required.
    \item[] Answer: \answerNA{} 
    \item[] Justification: The core method development in this research does not involve LLMs as any important, original, or non-standard components.

\end{enumerate}

\end{document}